\pdfoutput=1
\documentclass{article} 
\PassOptionsToPackage{numbers, compress}{natbib}
\usepackage[accsupp]{axessibility}

\usepackage{iclr2025_conference,times}
\usepackage{xspace}

\AtBeginDocument{
\newcommand{\dataset}{\textsc{MMSearch}\xspace}
\newcommand{\engine}{\textsc{MMSearch-Engine}\xspace}
}

\usepackage{amsmath,amsfonts,bm}

\def\eqref#1{equation~\ref{#1}}

\def\1{\bm{1}}

\DeclareMathAlphabet{\mathsfit}{\encodingdefault}{\sfdefault}{m}{sl}
\SetMathAlphabet{\mathsfit}{bold}{\encodingdefault}{\sfdefault}{bx}{n}

\usepackage{url}

\usepackage[nodayofweek]{datetime}
\usepackage{enumitem}
\usepackage{graphicx} 
\usepackage{booktabs} 
\usepackage{adjustbox}
\makeatletter
  \newcommand\figcaption{\def\@captype{figure}\caption}
  \newcommand\tabcaption{\def\@captype{table}\caption}
\makeatother
\usepackage{xcolor}    
\usepackage{adjustbox}
\usepackage{array}
\usepackage{multirow}
\usepackage{colortbl}
\usepackage{makecell}
\usepackage{caption}
\usepackage{textcomp}
\usepackage{url}
\usepackage{wrapfig}

\usepackage{xcolor}         
\usepackage{color, colortbl}
\definecolor{citecolor}{HTML}{2980b9}
\definecolor{linkcolor}{HTML}{c0392b}

\usepackage[hidelinks,breaklinks=true,colorlinks,bookmarks=false,citecolor=citecolor,linkcolor=linkcolor]{hyperref}

\definecolor{backred}{RGB}{255, 190, 190}
\definecolor{backblue}{RGB}{210, 230, 250}
\definecolor{verylightgray}{gray}{0.95} 

\title{
\begin{minipage}{.09\textwidth}
\centering
\includegraphics[width=1.1\linewidth]{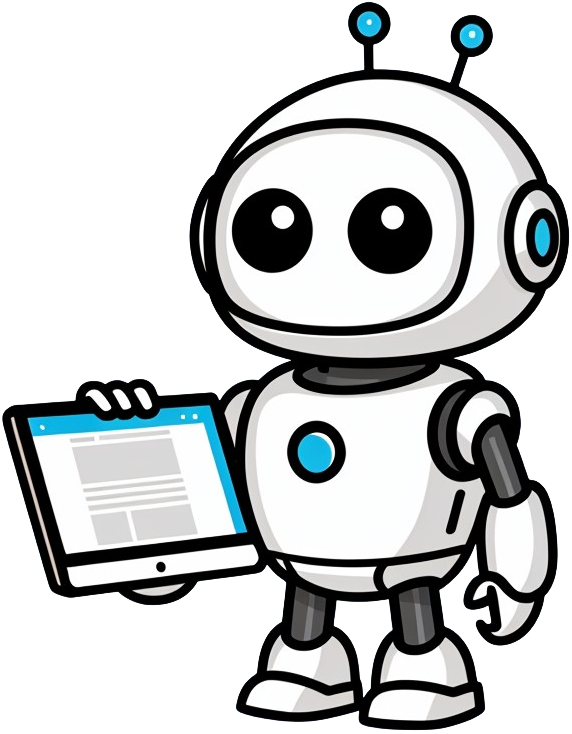}
\end{minipage}%
\hspace{0.45cm}
\begin{minipage}{.85\textwidth}
\dataset:\\Unveiling the Potential of Large Models as Multi-modal Search Engines
\end{minipage}}

\author{Dongzhi Jiang${^{1*}}$,
  Renrui Zhang${^{1,3 *\dagger}}$,
  Ziyu Guo$^{2}$,
  Yanmin Wu$^{5}$,
  Jiayi Lei$^{4}$\vspace{0.1cm} \\
  \textbf{Pengshuo Qiu}$^{4}$,
  \textbf{Pan Lu}$^{6}$, 
  \textbf{Zehui Chen}$^{3}$,
  \textbf{Chaoyou Fu}$^{7}$, 
  \textbf{Guanglu Song}$^{8}$\vspace{0.1cm} \\
  \textbf{Peng Gao}$^{4}$,
  \textbf{Yu Liu}$^{8}$, 
  \textbf{Chunyuan Li}$^{3}$,
  \textbf{Hongsheng Li}$^{1\ddagger}$\vspace{0.3cm} \\
  $^1$CUHK MMLab\hspace{0.1cm} \&\hspace{0.05cm} $^2$MiuLar Lab  \quad  $^3$ByteDance \quad $^4$Shanghai AI Laboratory \\
   $^5$Peking University \quad $^6$Stanford University \quad $^7$Nanjing University \quad$^8$Sensetime Research\vspace{0.1cm}\\
  \texttt{\{dzjiang,renruizhang,ziyuguo\}@link.cuhk.edu.hk} \vspace{1em}  \\
\vspace{0.5cm} $^*$ Equal contribution \ \ $^{\dagger}$ Project lead \ \ $^{\ddagger}$ Corresponding author \\
\hspace{0.2cm}Project Page: \url{https://mmsearch.github.io}
}

\iclrfinalcopy 
\begin{document}

\maketitle



\begin{abstract}
The advent of Large Language Models (LLMs) has paved the way for AI search engines, e.g., SearchGPT, showcasing a new paradigm in human-internet interaction.
However, most current AI search engines are limited to text-only settings, neglecting the multimodal user queries and the text-image interleaved nature of website information. 
Recently, Large Multimodal Models (LMMs) have made impressive strides. Yet, whether they can function as AI search engines remains under-explored, leaving the potential of LMMs in multimodal search an open question.
To this end, we design the first multimodal AI search engine pipeline, \textbf{\engine}, to empower any LMMs with multimodal search capabilities. On top of this, we introduce \textbf{\dataset}, a comprehensive evaluation benchmark to assess the multimodal search performance of LMMs. The curated dataset contains 300 manually collected instances spanning 14 subfields, which involves no overlap with the current LMMs' training data, ensuring the correct answer can only be obtained within searching. By using \engine, the LMMs are evaluated by performing three individual tasks (requery, rerank, and summarization), and one challenging end-to-end task with a complete searching process.
We conduct extensive experiments on closed-source and open-source LMMs. Among all tested models, GPT-4o with \engine achieves the best results, which surpasses the commercial product, Perplexity Pro, in the end-to-end task, demonstrating the effectiveness of our proposed pipeline. 
We further present error analysis to unveil current LMMs still struggle to fully grasp the multimodal search tasks, and ablation study to indicate the potential of scaling test-time computation for AI search engine. We hope \dataset~may provide unique insights to guide the future development of multimodal AI search engine.
\end{abstract}

\begin{figure}[ht]
\centering
\includegraphics[width=\textwidth]{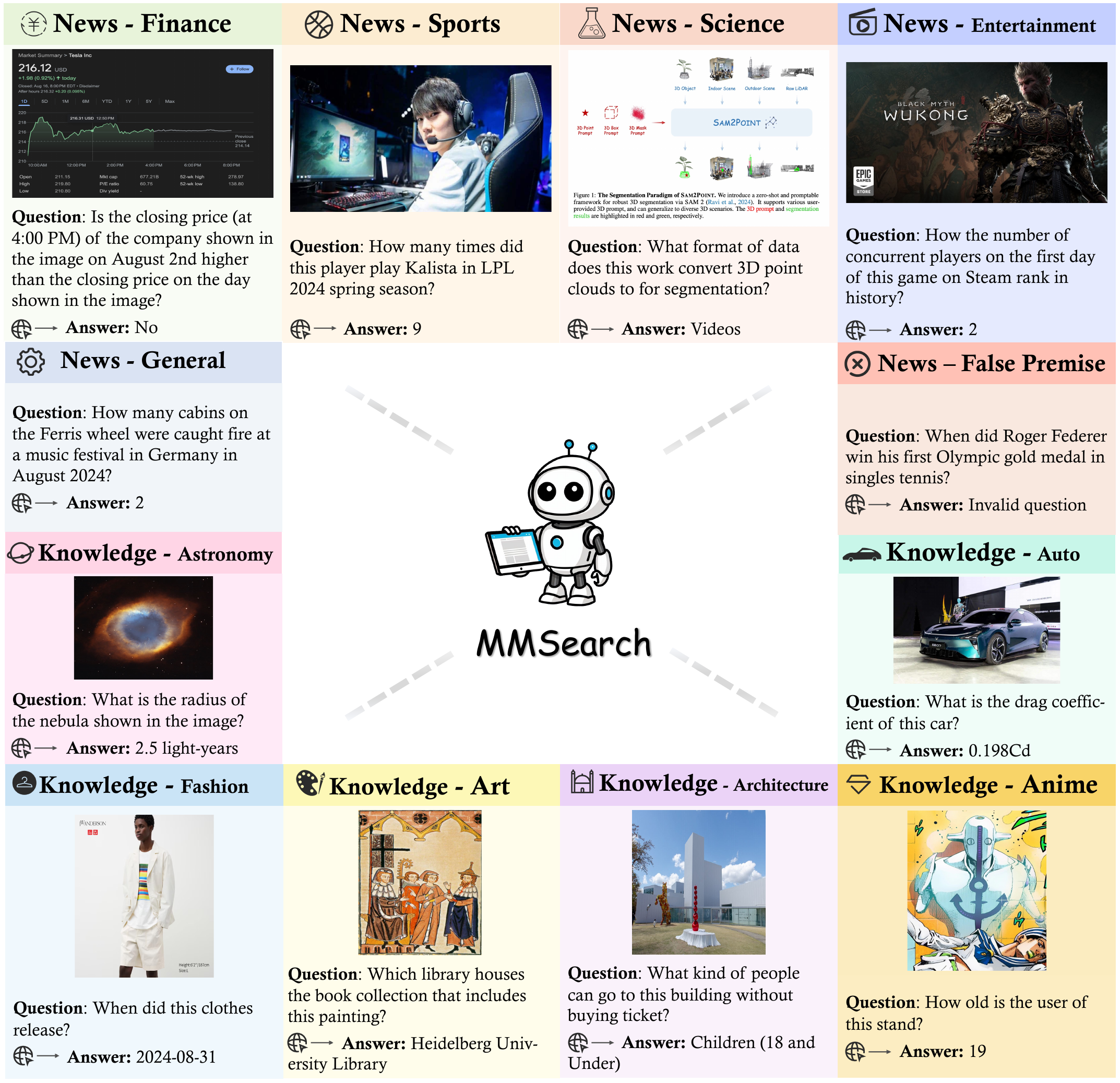} 
\caption{\textbf{Overview of the \dataset Benchmark.} \dataset aims to evaluate any LMM's potential to be a multimodal AI search engine. The benchmark contains two primary areas: latest news and rare knowledge to ensure no overlap with LMM's inherent knowledge.}
\label{fig:teaser}
\end{figure}

\vspace{0.2cm}
\section{Introduction}
Search engines~\citep{brin1998anatomy} have been the main tools for humans to navigate through the overwhelming quantity of online resources. Recently, Large Language Models (LLMs)~\citep{OpenAI2023ChatGPT, OpenAI2023GPT4TR, touvron2023llama} have demonstrated impressive performance on various zero-shot downstream applications. On top of this, AI search engine~\citep{searchgpt}, which integrates LLMs with traditional search engines, stands among one of the most promising ones. It points the direction of the next-generation interaction paradigm of human and Internet. Combining the language understanding ability of LLMs and up-to-date information from the Internet, AI search engines could better grasp the user's intention and summarize contextual-aligned answers from the raw web information.
These systems can only process textual queries and interpret textual web content, significantly constraining user query scenarios and information-seeking methods~\citep{barbany2024leveraging, xie2024large}. This limitation impacts both the range of input queries and the accuracy of results~\citep{jiang2024multi, chen2021websrc, lu2024weblinx}, particularly given the complexity and interleaved nature of modern websites~\citep{liu2024visualwebbench}.
For example, consider a scenario where you possess numerous medals belonging to your grandfather but are unaware of their specific names. A multimodal AI search engine could match photographs of these medals with an interleaved table of images and text retrieved from the Internet, thereby identifying each medal. In contrast, text-only search engines can neither take photographs for searching nor understand the interleaved table.
Hence, a multimodal AI search engine is crucial for advancing information retrieval and analysis. 

On the other hand, with the recent rapid advancements, Large Multimodal Models (LMMs)~\citep{liu2023llava, lin2023sphinx, openai2023gpt4v,gao2024sphinx,zhang2024llamaadapter} have showcased significant abilities across diverse scenarios, including general image understanding~\citep{fu2023mme,liu2023mmbench, Yu2023MMVetEL}, expert image reasoning~\citep{zhang2024mavis, gao2023g,zhang2024mathverse,sciverse}, multi-image perception~\citep{li2024llava-str, wang2024muirbench, jiang2024mantis,li2024llava-inter}, and spatial environment perception~\citep{guo2023point,yang2023lidar,han2023imagebind}. Despite these developments, a framework for LMMs to function as multimodal AI search engines remains largely unexplored. Consequently, the potential of LMMs in multimodal searching also remains a significant open question.

To bridge this gap, we first present \engine, a multimodal AI search engine pipeline, empowering any LMMs with advanced search capabilities. \engine~ maximizes the utilization of LMMs' multimodal information comprehension abilities, incorporating both visual and textual website content as information sources. On top of this, we introduce \dataset, a multimodal AI search engine benchmark to comprehensively evaluate LMMs' searching performance. The design of \engine~facilitates the zero-shot evaluation of any LMMs within the context of AI search engine. Our experiment covers state-of-the-art closed-source~\citep{openai2023gpt4v, anthropic2024claude35, team2023gemini} and open-source LMMs~\citep{li2024llava-ov, Qwen2-VL, chen2024far, ye2024mplug}. Our efforts are summarized as follows:

\begin{enumerate}[label=\roman*.]
    \item \textit{\textbf{\engine, the first multimodal AI search engine pipeline for any LMMs,}}
    empowering large models for multimodal searching. In contrast with the conventional text-only AI search engines, \engine~fully integrates multimodal information in two ways: (i) for queries containing images, we conduct web searches across both textual and visual modalities. We utilize Google Lens~\citep{lens} to identify critical visual information from the input image; (ii) all search results are presented in both textual and visual formats, ensuring a comprehensive understanding of the interleaved website content.
    The working flow of \engine~contains multi-round interaction between LMM and the Internet. 
    The LMM needs to first \textit{requery} the user question into a search-engine-friendly format. Then, the LMM \textit{reranks} the retrieved websites based on its helpfulness. Finally, the LMM is required to \textit{summarize} the answer based on the most informative webpage content selected from the rerank. Thanks to the design of the pipeline, we propose a step-wise evaluation strategy on the three core tasks within the searching process: \textit{requery}, \textit{rerank}, and \textit{summarization}. The final score is weighted by the end-to-end evaluation results and scores of the three core tasks.
    \vspace{0.15cm}
    
    \item \textit{\textbf{\dataset, a comprehensive benchmark for multimodal AI search engines,}} 
    which, to our best knowledge, serves as the first evaluation dataset to measure LMMs' multimodal searching capabilities. Our benchmark categorizes searching queries into two primary areas: \textit{News} and \textit{Knowledge}, as shown in Fig.~\ref{fig:teaser}. 
    We employ different strategies for these two areas to ensure the challenging nature of the benchmark. 
    \textit{News} area covers the latest news at the time of data collection (August, 2024). This is to guarantee the answers to the queries will not be present in the training data of LMMs. 
    As for the area of \textit{Knowledge}, we collect queries requiring rare knowledge and then select the queries unable to be answered by current SoTA LMMs such as GPT-4o~\citep{openai2024gpt4o} or Claude-3.5~\citep{anthropic2024claude35}. The two areas sum up to 14 subfields. In total, \dataset~encompasses 300 meticulously collected queries, with 2901 unique images. 
    \vspace{0.15cm}

    \item \textit{\textbf{Extensive experiments and error analysis for future development direction.}}
    We evaluate popular closed-source models and open-source LMMs on \dataset. GPT-4o achieves the best overall performance across different tasks. Surprisingly, our \engine~equipeed with SoTA LMMs, such as GPT-4o and Claude 3.5 Sonnet, even surpasses the prominent commercial AI search engine Perplexity Pro~\citep{perplexity} in the end-to-end task. 
    Our thorough error analysis reveals that current LMMs still struggle to generalize to multimodal search-specific tasks. Their poor requery and rerank capabilities significantly limit their ability to correctly identify useful websites and extract relevant answers. Additionally, we identify five error types for requery and summarization tasks, respectively. We find that current LMMs cannot fully understand the requery task and do not know how to query the search engine. As for the summarization task, LMMs often have difficulty in extracting useful information, either from text or images. These capabilities are essential for LMMs to function as robust multimodal search engines and require further development. We also conduct a preliminary ablation study to explore the potential of scaling test-time computation versus scaling model size~\citep{gpt4o1}. Initial results indicate that scaling test-time computation demonstrates superior performance in this task.

\end{enumerate}
\clearpage

\begin{figure}[t]
\centering
\includegraphics[width=\textwidth]{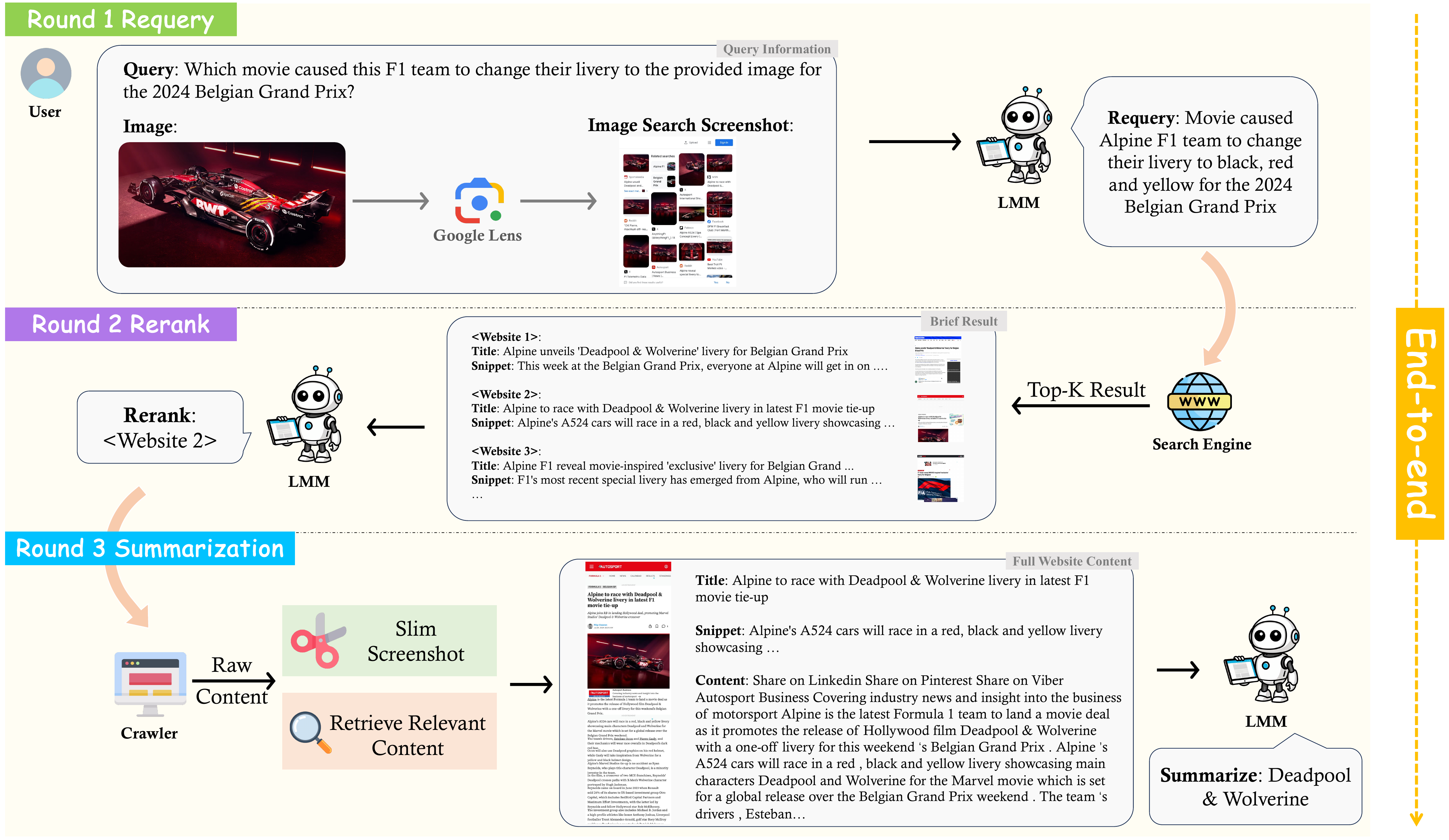} 
\caption{\textbf{Pipeline of \engine.} The process comprises three sequential stages executed by a LMM: (i) requery, (ii) rerank, and (iii) summarization. In the end-to-end evaluation task, the LMM completes these three stages sequentially to generate the final output.}
\label{fig:pipeline}
\end{figure}

\section{\dataset}
In Section~\ref{sec2-pipeline}, we first detail the design of our multimodal AI search engine pipeline, which serves as both data collection and evaluation tools. Then, in Section~\ref{sec2-benchmark}, we detail the data composition and collection of the curated multimodal search benchmark \dataset. Then, in Section~\ref{sec2-evaluation}, we elaborate on our step-wise evaluation strategy. Finally, we detail the dynamic nature of our benchmark in Section~\ref{sec2:benchmark_evolution}.

\subsection{\engine: A Multimodal AI Search Engine Pipeline}
\label{sec2-pipeline}
The searching process is a complex action including multi-round interactions between LMMs and conventional search engines. We develop a delicate pipeline that queries LMMs multiple times to accomplish this task. Leveraging the image comprehension capabilities of LMMs, we incorporate two types of visual data. First, we incorporate Google Lens~\citep{lens} to search for information from the image. The second type of visual data is the screenshot of the retrieved websites, in the purpose of preserving the original format of website content. Our framework is shown in Fig.~\ref{fig:pipeline}. Below we detail how an LMM works with this pipeline, which comprises three sequential phases:

\begin{enumerate}[label=\roman*.]
    \item \textit{Requery.} 
    The query direct from users may contain references to certain information in the image, e.g., the \textit{News-Finance} example shown in Fig.~\ref{fig:teaser}. Since a conventional search engine only accept text-only input, it is necessary for LMM to translate the image content and combine it with the query to ask a valid question to it. In addition, the raw user query may be ambiguous or inefficient sometimes~\citep{chan2024rq, ma2023query}, reformulating the query to be more clear is also a must for LMM. If the user query contains an image, we incorporate the screenshot of the image search result from the google lens~\citep{lens}. We treat the user query, user image, and the image search screenshot as basic information of the query. This information will be input to LMM in every round in the pipeline. For the requery round, we prompt LMM to output a requery to a conventional search engine.
    \vspace{0.1cm}
    \item \textit{Rerank.} 
    The requery is sent to a search engine API, e.g., DuckDuckGo, to retrieve top $K$ relevant websites. Depending on the requery quality, not all retrieved websites are necessarily relevant for query answering. Hence, we prompt LMM to select the most informative website for answer summarization. Due to the LMM's context length limitations and the extensive content of websites, we provide only essential information of each website, which we term \textit{brief results}. These brief results include the title, the snippet, and a screenshot of the webpage's top section, which serves as the input for LMM's reranking. The inclusion of the screenshot serves two purposes. First, the screenshot offers a visual cue to assess the web's credibility, as a well-organized website often appears more trustworthy than one cluttered with advertisements~\citep{fogg2001makes, sillence2004trust}. Additionally, the screenshot may contain essential visual information. For instance, it might include images similar or identical to query images, as shown in the \textit{Website 2} in Fig.~\ref{fig:pipeline}.
    \vspace{0.1cm}
    \item \textit{Summarization.} 
    We start by crawling the selected website to gather all the available information. We parse the HTML to obtain the raw textual content and capture a full-page screenshot of the website. However, there are two issues: the raw content tends to be extensively lengthy and disorganized, while substantial areas in the full-page screenshot are blank due to the advertisement blocks on the website. These two issues lead to a large number of input tokens filled with irrelevant information. 
    To enhance data efficiency, we slim the screenshot and retrieve the relevant content before inputting them to LMM. 
    For the full-page screenshot, we identify the blank areas and remove them iteratively, detailed in Appendix~\ref{appendix:exp_details}.
    As for the text content, we apply a text embedding model~\citep{bge-m3} to retrieve a maximum of 2K tokens relevant to the requery from the raw content. We define the slimmed screenshot and the retrieved content as \textit{full website content}. Finally, we input the full website content, website title, and website snippet, along with the query information, to LMM for summarizing the answer.

\end{enumerate}



\begin{figure*}[t]
\centering
\begin{minipage}[c]{0.45\textwidth}
\small
\centering

  \tabcaption{\textbf{Key Statistics of \dataset.}}
  \label{t1}
  \centering
  \begin{adjustbox}{width=\linewidth}
   \begin{tabular}{lc}
 \toprule
 \textbf{Statistic} & \textbf{Number} \\
 \midrule
  Total questions & 300 \\
  ~- Questions with images & 171 (57.0\%) \\
  ~- Questions without images & 129 (43.0\%) \\
  Total Websites & 2,280 \\
  Total Areas/Subfields & 2/14 \\
  \midrule
 Number of unique images & 2,901 \\
 ~- Query images & 163 \\
 ~- Google search images & 163 \\
 ~- Top section screenshot images & 2,280 \\
 ~- Full-page screenshot images & 295 \\
 Number of unique questions & 289 \\
 Number of unique requeries & 289 \\
 Number of unique reranked websites & 2,400 \\
 Number of unique answers & 264 \\
 \midrule
 Maximum question length & 41 \\
 Maximum answer length & 12 \\
 Average question length & 14.0 \\
 Average answer length & 1.9 \\
 \bottomrule
 \end{tabular}
 \end{adjustbox}
 \label{table:statistics}
\end{minipage}
\qquad
\begin{minipage}[c]{0.42\textwidth}
\centering
\vspace{-0.2cm}
\caption{\textbf{Area and Subfield Distribution of \dataset.}}
\label{fig:pie}
\vspace{0.15cm}
\includegraphics[width=1.05\linewidth]{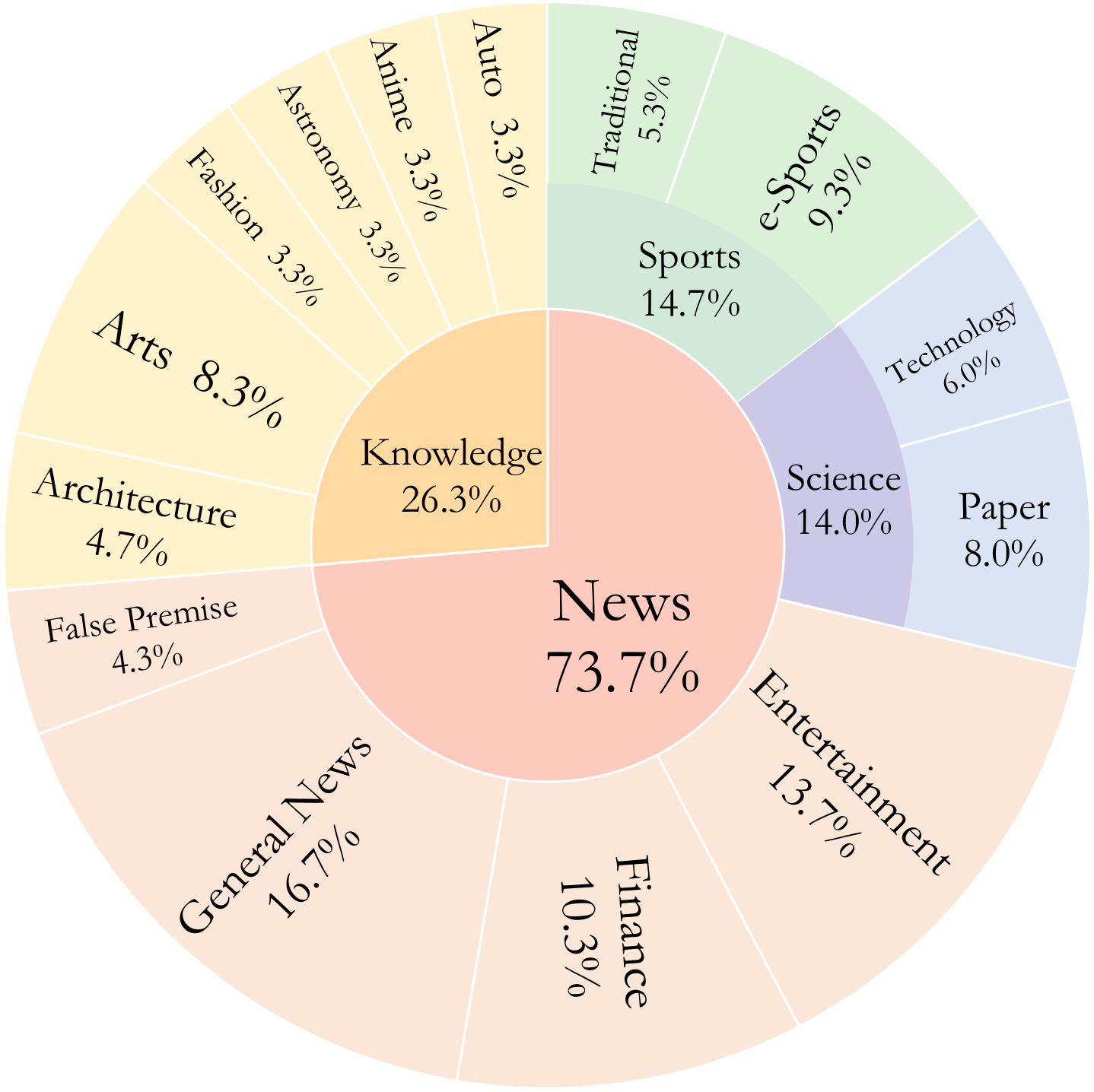}
\end{minipage}
\label{statistics_and_pie}
\end{figure*}

\subsection{Data Composition and Collection}
\label{sec2-benchmark}
To thoroughly assess multimodal search proficiency, we compile a comprehensive problem set
covering a broad spectrum of news topics, specialized knowledge domains, and query image patterns.
This widespread collection for \dataset~aims to simulate diverse user searching scenarios, ensuring a robust evaluation of LMMs' capabilities in multimodal search.

\textbf{Data Composition and Categorization.} 
Our benchmark aims to isolate LMMs' inherent knowledge and assess their actual search capabilities. We focus on two primary areas: \textit{News} and \textit{Knowledge}. For the \textit{News} area, the queries are related to the latest news at the time of data collection (August 2024). This guarantees no overlap between the current LMMs' training data and questions in our benchmark. All questions in this area are recorded with their occurrence time. For fairness, LMMs with recently updated knowledge should be tested on queries that occurred after their latest data update. Due to its time-sensitive nature, the \textit{News} area serves as a dynamic part of our benchmark. Please refer to Section \ref{sec2:benchmark_evolution} for details. As for the \textit{Knowledege} area, we focus on rare knowledge in targeted domains. Each question proposed by an annotator is verified to be beyond the capabilities of state-of-the-art LLMs such as GPT-4o~\citep{openai2024gpt4o} or Claude 3.5 Sonnet~\citep{anthropic2024claude35}. The \textit{Knowledge} area serves as a static component of our benchmark and remains constant over time. We collect a total of 300 queries across the 2 primary areas and 14 subfields. Detailed statistics for data composition and categorization are presented in Table~\ref{table:statistics} and Fig.~\ref{fig:pie}. Definitions of each subfield are illustrated in Appendix~\ref{appendix:data_details_subfield}.


\textbf{Data Collection and Review Process.}
Thanks to the design of our pipeline, the data collection process follows similar procedure introduced in the pipeline. An annotator is first required to propose a query and provide its answer, either sourced from the latest news or rare knowledge. The annotator then formulates a requery based on the query information. After $K$ websites are retrieved from the search engine, the annotator is required to divide all $K$ websites into three sets based on the brief results: \textit{valid} (likely to contain the answer), \textit{unsure} (relevance is difficult to determine), and \textit{invalid} (entirely irrelevant to the question). We mandate that at least one website must be classified as \textit{valid}; if this criterion is not met, the annotator is required to adjust the requery to obtain new search results. Finally, we randomly pick one website from the \textit{valid} set and obtain its full content. To ensure the question is answerable, another annotator is employed to give an answer to the query based on the full content. If the answer is incorrect, the question needs to be revised until it is answerable. 

\begin{figure}[!t]
\centering
\includegraphics[width=\textwidth]{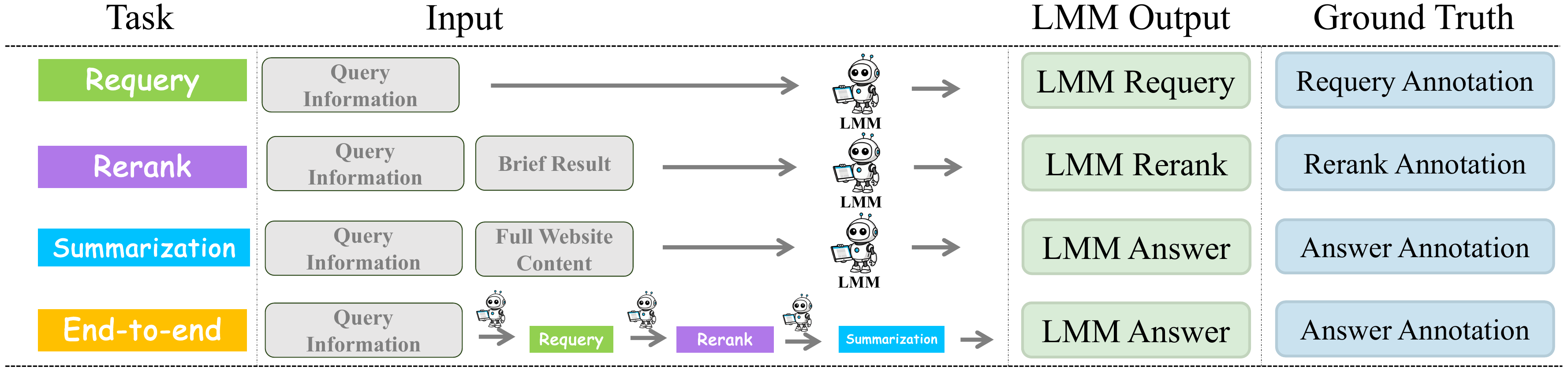} 
\caption{\textbf{Outline of Evaluation Tasks, Inputs, and Outputs.} Our evaluation contains four tasks. The requery, rerank, and summarization tasks assess the LMM's proficiency in individual pipeline rounds. The end-to-end task simulates a real-world search scenario by sequentially executing all three stages. An example of the input and output is shown in Fig.~\ref{fig:mainpaper_example}.}
\label{fig:evaluation_strategy}
\vspace{-0.2cm}
\end{figure}

\subsection{Evaluation Protocol}
\label{sec2-evaluation}

In contrast with previous LMM benchmarks, the multimodal search process of LMM contains multiple rounds. Only the end-to-end evaluation of the final answer is inadequate to reveal the models' deficiency in each core searching step. For example, the errors made by the model may occur during the summarization process, but it might also stem from choosing an incorrect website during the reranking stage.
To this end, we propose a step-wise strategy to evaluate the LMMs' capability on the three core searching steps, in addition to the end-to-end evaluation.

\vspace{-0.2cm}

\begin{itemize}
    \item \textbf{End-to-end score ($\mathbf{S}_{e2e}$):} We compute the F1 score between the predicted answer and the ground truth to judge if the answer is correct.
    \item \textbf{Requery score ($\mathbf{S}_{req}$):} We apply the average of ROUGE-L and BLEU-1 scores to measure the similarity between the model's requery and human-annotated requery.
    \item \textbf{Rerank score ($\mathbf{S}_{rer}$):} The rerank score is derived from the LMM's selection among $K$ pre-defined websites. The score values is 1.0 for valid set, 0.5 for unsure set, and 0 for invalid set or incorrect format.
    \item \textbf{Summarization score ($\mathbf{S}_{sum}$):} Again, we compute the F1 score of LMM's answer based on a pre-defined website content against ground truth.
\end{itemize}

\vspace{-0.2cm}

The input, output, and ground truth of the four tasks are visualized in Fig.~\ref{fig:evaluation_strategy}. The final score is weighted by these four scores. We assign the highest weight (75\%) to the end-to-end task, as it reflects the real-world multimodal search capability. The remaining 25\% is distributed among the intermediate steps: 10\% each for the rerank and summarization tasks, and 5\% for the requery task. The lower weight for the requery task accounts for the inherent uncertainty in this process. The scoring process can be formulated as:

\vspace{-0.1cm}
\begin{equation}
    \mathbf{S}_{final} =  0.75\cdot\mathbf{S}_{e2e} + 0.05\cdot\mathbf{S}_{req} + 0.1\cdot\mathbf{S}_{rer} + 0.1\cdot\mathbf{S}_{sum} 
\end{equation}

\subsection{Benchmark Evolution}
\label{sec2:benchmark_evolution}
\begin{figure}[!h]
\centering
\vspace{-0.1cm}
\includegraphics[width=\textwidth]{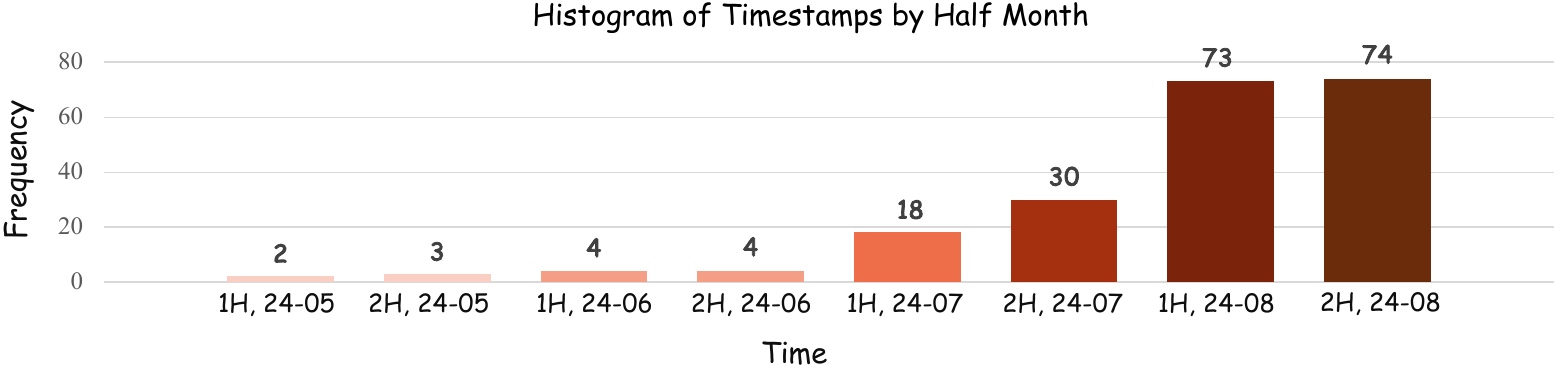} 
\caption{\textbf{Timestamps Distribution of Questions in the \textit{News} Area.} All events of our collected data occurred after May 2024. The majority of data concentrates on August. This ensures the data captures only recent events, falling beyond the knowledge cutoff dates of LMMs. The False Premise subfield is not included, since it is infeasible to determine the timestamp for an event that never occurred.
}
\label{fig:time_distribution}
\end{figure}

In Fig.~\ref{fig:time_distribution}, we showcase the statistics of data timestamp distribution in the \textit{News} area. Our dataset spans from 1st May 2024 to 31th August 2024. By the time of evaluation, we inspect the knowledge cutoff dates of the closed-source models. Claude 3.5 Sonnet reports a knowledge cutoff of April 2024, while both GPT-4V and GPT-4o state they lack information from 2024. For open-source models, we examine their release dates and training data, confirming that none possess knowledge beyond May 2024. This temporal gap ensures the fairness of our evaluation, as the models' performance solely reflects their multimodal search capabilities rather than pre-existing knowledge.
We will update the \textit{News} area if a new LMM's training data may overlap with our collection period. 

\begin{figure}[p]
\centering
\includegraphics[width=\textwidth]{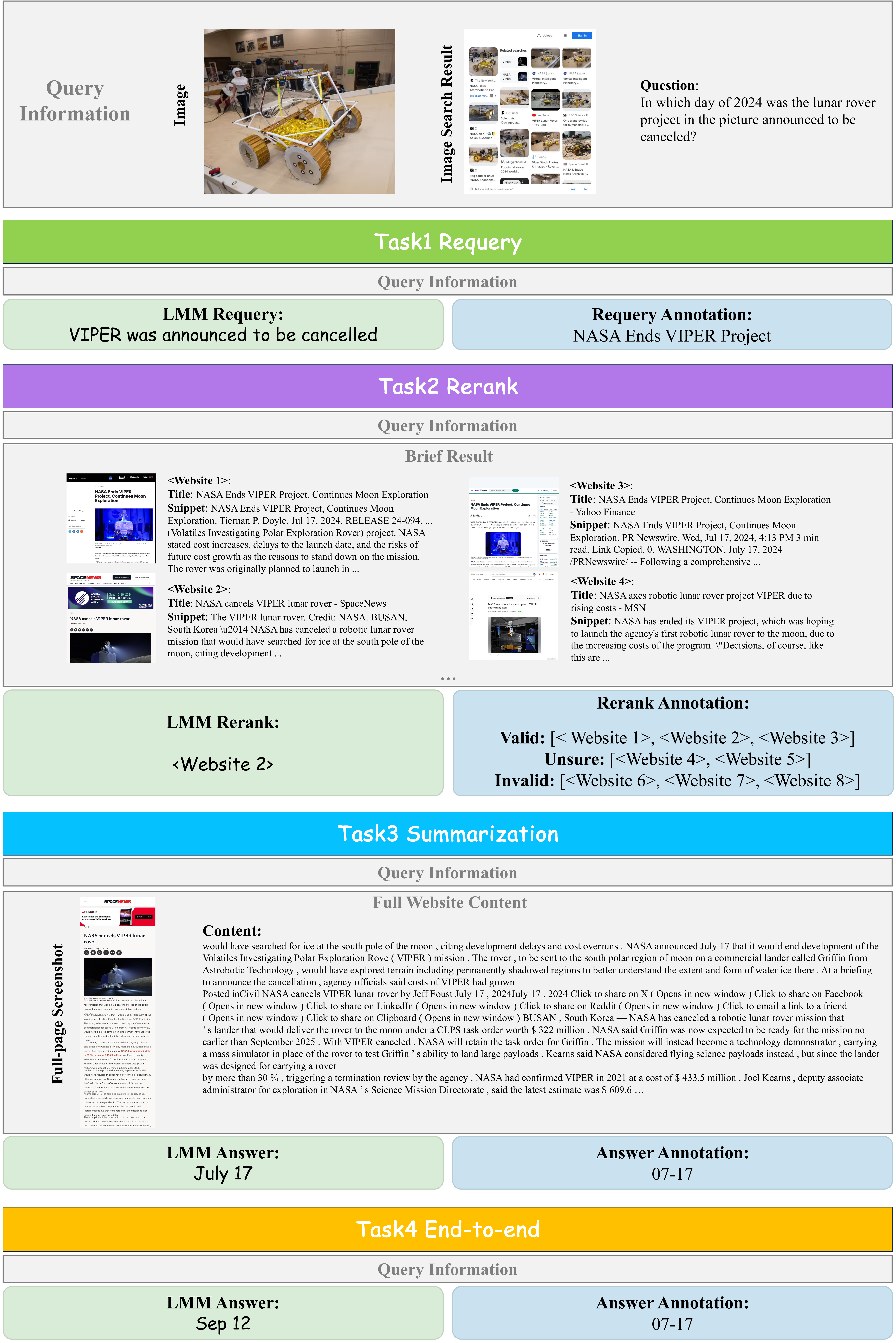} 
\caption{\textbf{Example Input, LMM Output, and Ground Truth for Four Evaluation Tasks.} The color-coding of each module corresponds to Fig.~\ref{fig:evaluation_strategy}. Task1 Requery (green), Task2 Rerank (purple), Task3 Summarization (blue), and Task4 End-to-end (yellow) are shown. Image best viewed in color.}
\label{fig:mainpaper_example}
\end{figure}

\section{Experiment}
In this section, we conduct a systematic evaluation of existing LMMs on \dataset. We first introduce the experimental setup in Section~\ref{sec3-experiment_setup}. Then, we detail the quantitative results in Section~\ref{sec3-experiment_analysis} and narrate the error analysis in Section~\ref{sec3-error_analysis}. Finally, we explore scaling test-time compute versus scaling model size in Section~\ref{sec3-ttc}.

\begin{table*}[!t]
\centering
\caption{\textbf{Evaluation Results of Four Tasks in \dataset.} We report the scores of news and knowledge areas and their average score in each task. Subscript \textit{AnyRes} indicates original resolution image input; otherwise, low-resolution images were used. The highest scores for \colorbox{backred!50}{closed-source} and \colorbox{backblue!75}{open-source} LMMs are marked in red and blue. \textit{For open-source LMMs, we adopt the models with 7B parameters unless otherwise specified.}}
 \renewcommand\tabcolsep{2.5pt}
 \renewcommand\arraystretch{1.25}
 \resizebox{1.0\linewidth}{!}{
    \begin{tabular}{l|>{\columncolor{verylightgray}}c|cc|>{\columncolor{verylightgray}}c|cc|>{\columncolor{verylightgray}}c|cc|>{\columncolor{verylightgray}}c|cc|>{\columncolor{verylightgray}}c|cc}
    \toprule
    \multirow{2}*{\makecell*[l]{\large Model}} &\multicolumn{3}{c|}{All}
    &\multicolumn{3}{c|}{End-to-end} 
    &\multicolumn{3}{c|}{Requery}
    &\multicolumn{3}{c|}{Rerank}
    &\multicolumn{3}{c}{Summarize}\\
    \cmidrule{2-16}
    & Avg \ & News & \ Know.  & Avg & News& \ Know.& Avg \ & News& \ Know.& Avg \ & News& \ Know.& Avg \ & News & \ Know. \\
    \midrule
    \multicolumn{16}{c}{\textit{Baselines}}\\
    \cmidrule{1-16}
    Human & 69.2 & 69.6 & 68.1 & 68.2 & 68.6 & 67.1 & 43.7 & 45.0 & 40.1 & 85.7 & 87.3 & 81.2 & 72.8 & 71.4 & 76.7 \\
    \cmidrule{1-16}
    \multicolumn{16}{c}{\textit{Commercial AI Search Engines}}\\
    \cmidrule{1-16}
    Perplexity Pro~\citep{perplexity} & - & - & - &47.8 & 52.7 & 34.1 - & - & - & - & - & - & - & - & - & - \\
    \cmidrule{1-16}
    \multicolumn{16}{c}{\textit{Closed-source LMMs~~ with \engine}}\\
    \cmidrule{1-16}
    Claude 3.5 Sonnet~\citep{anthropic2024claude35} & 53.5 & 53.1 & 54.7 & 49.9 & 49.3 & 51.6 & 42.0 & 43.6 & 37.7 & 80.2 & 78.7 & 84.2 & 59.4 & 60.3 & 57.0 \\
    GPT-4V~\citep{openai2023gpt4v} & 55.0 & 55.0 & 55.3 & 52.1 & 52.2 & 51.9 & 45.7 & 49.2 & 35.8 & 79.3 & 76.9 & 86.1 & 57.4 & 56.7 & 59.4 \\
    GPT-4o~\citep{openai2024gpt4o} & \colorbox{backred!50}{62.3} & \colorbox{backred!50}{61.2} & \colorbox{backred!50}{65.3} & \colorbox{backred!50}{60.4} & \colorbox{backred!50}{59.0} & \colorbox{backred!50}{64.5} & \colorbox{backred!50}{46.8} & \colorbox{backred!50}{49.9} & \colorbox{backred!50}{38.0} & \colorbox{backred!50}{83.0} & \colorbox{backred!50}{82.4} & \colorbox{backred!50}{84.8} & \colorbox{backred!50}{63.1} & \colorbox{backred!50}{62.2} & \colorbox{backred!50}{65.6} \\
    \cmidrule{1-16}
    \multicolumn{16}{c}{\textit{Open-source LMMs~~ with \engine}}\\
    \cmidrule{1-16}
    Mantis~\citep{jiang2024mantis} & 18.7 & 19.8 & 15.9 & 15.8 & 16.4 & 14.3 & 20.1 & 24.6 & 7.4 & 39.7 & 41.0 & 36.1 & 19.2 & 22.0 & 11.5 \\
    InternLM-XC2.5~\citep{zhang2024internlm} & 22.2 & 22.8 & 20.5 & 22.9 & 23.6 & 20.8 & 25.0 & 24.3 & 27.0 & 0.0 & 0.0 & 0.0 & 37.7 & 38.6 & 35.0 \\
    InternLM-XC2.5\textsubscript{\textit{AnyRes}} & 22.3 & 23.9 & 17.5 & 23.2 & 25.4 & 16.9 & 21.7 & 19.8 & 26.9 & 0.0 & 0.0 & 0.0 & 37.7 & 38.6 & 35.1 \\
    LLaVA-NeXT-Interleave~\citep{li2024llava-inter} & 28.3 & 29.2 & 25.6 & 23.0 & 23.8 & 20.5 & 26.2 & 30.7 & 13.5 & 55.3 & 58.6 & 46.2 & 42.5 & 40.0 & 49.3 \\
    mPlug-Owl3~\citep{ye2024mplug} & 32.1 & 34.8 & 24.4 & 24.6 & 28.1 & 14.9 & 32.6 & 36.7 & 21.2 & 74.3 & 73.5 & 76.6 & 45.6 & 45.6 & 45.4 \\
    mPlug-Owl3\textsubscript{\textit{AnyRes}} & 33.9 & 35.5 & 29.3 & 27.3 & 29.4 & 21.2 & 31.8 & 36.1 & 19.9 & 74.5 & 72.9 & 79.1 & 43.9 & 43.6 & 44.6 \\
    InternVL2~\citep{chen2024far} & 34.3 & 35.7 & 30.2 & 30.9 & 32.5 & 26.2 & 32.3 & 36.1 & 21.6 & 46.5 & 49.3 & 38.6 & 48.5 & 45.9 & 55.8 \\
    InternVL2\textsubscript{\textit{AnyRes}} & 34.1 & 34.2 & 33.7 & 30.0 & 30.3 & 29.0 & 31.4 & 35.5 & 19.7 & 53.2 & 52.3 & 55.7 & 46.9 & 44.4 & 53.7 \\
    Idefics3~\citep{laurençon2024building} & 36.2 & 38.5 & 29.6 & 29.3 & 32.3 & 20.8 & 31.0 & 37.3 & 13.5 & 76.5 & 73.3 & 85.4 & 50.3 & 51.1 & 48.1 \\
    Idefics3\textsubscript{\textit{AnyRes}} & 35.7 & 38.2 & 28.7 & 30.1 & 32.9 & 22.3 & 27.2 & 32.2 & 13.2 & 72.7 & 73.1 & 71.5 & 45.2 & 46.3 & 42.1 \\
    LLaVA-OneVision~\citep{li2024llava-ov} & 36.6 & 39.4 & 28.9 & 29.6 & 33.1 & 19.7 & 35.8 & 40.3 & 23.2 & 72.8 & 73.5 & 70.9 & 53.5 & 51.8 & 58.5 \\
    Qwen2-VL\textsubscript{\textit{AnyRes}}~\citep{Qwen2-VL} & 45.3 & 44.1 & 48.7 & 40.3 & 39.2 & 43.5 & {39.0} &{41.9} & {30.8} & 76.7 & 73.8 & 84.8 & 54.7 & 52.7 & 60.4 \\
    LLaVA-OneVision~(72B) & {50.1} & {50.1} & {50.2} & {44.9} & {45.1} & {44.1} & 42.9 & 45.9 & 34.3 & \colorbox{backblue!75}{82.2} & \colorbox{backblue!75}{80.5} & 86.7 & \colorbox{backblue!75}{61.4} & \colorbox{backblue!75}{59.1} & \colorbox{backblue!75}{67.7} \\
    Qwen2-VL\textsubscript{\textit{AnyRes}}~(72B) & \colorbox{backblue!75}{52.7} & \colorbox{backblue!75}{52.0} & \colorbox{backblue!75}{54.5} & \colorbox{backblue!75}{49.1} & \colorbox{backblue!75}{48.8} & \colorbox{backblue!75}{49.8} & \colorbox{backblue!75}{44.7} & \colorbox{backblue!75}{47.2} & \colorbox{backblue!75}{37.6} & {76.7} & {72.9} & \colorbox{backblue!75}{87.3} & {59.6} & {57.5} & {65.7} \\
    \bottomrule
    \end{tabular}
}
\label{table:main_result}
\end{table*}

\subsection{Experiment Setup}
\label{sec3-experiment_setup}
\textbf{Evaluation Models.}
We examine the performance of foundation models across three distinct categories on \dataset: (a) \textit{Commercial AI Search Engines}, represented by Perplexity~\citep{perplexity}. We test the pro version of Perplexity, which takes only the user query and image as input. Since SearchGPT~\citep{searchgpt} has not been public yet, we do not test on it. (b) \textit{Closed-source LMMs}, represented by models like GPT-4V~\citep{openai2023gpt4v}, GPT-4o~\citep{openai2024gpt4o}, and Claude 3.5 Sonnet~\citep{anthropic2024claude35}, and (c) \textit{Open-source LMMs}, featuring models such as LLaVA-OneVision-7B~\citep{li2024llava-ov} (Qwen2-7B~\citep{qwen2}), LLaVA-OneVision-72B~\citep{li2024llava-ov} (Qwen2-72B~\citep{qwen2}),  LLaVA-NeXT-Interleave~\citep{li2024llava-inter} (Qwen1.5-7B~\citep{qwen2}), InternVL2~\citep{chen2024far} (InternLM2.5-7B-Chat~\citep{cai2024internlm2}), InternLM-XC2.5~\citep{zhang2024internlm} (InternLM2-7B~\citep{cai2024internlm2}), Qwen2-VL-7B~\citep{Qwen2-VL} (Qwen2-7B~\citep{qwen2}), Qwen2-VL-72B~\citep{Qwen2-VL} (Qwen2-72B~\citep{qwen2}), mPlug-Owl3~\citep{ye2024mplug} (Qwen2-7B~\citep{qwen2}), Idefics3~\citep{laurençon2024building} (LLaMA3.1-7B-Instruct~\citep{llama3modelcard}), and Mantis~\citep{jiang2024mantis} (LLaMA3-7B~\citep{llama3modelcard}). Note that the open-source LMMs' sizes are 7B unless otherwise specified.

\paragraph{\textbf{Implementation Details.}}
We set the number of retrieved websites $K$ as 8.
All our experiments of open-source models are conducted without any fine-tuning on search data or tasks. As for the prompts, the requery prompt contains 3 examples to better guide LMMs to output a valid requery, while prompts for other tasks use zero-shot settings.  We prompt the LMM to output as few words as possible for a better match with the ground truth. 
We employ the metric introduced in Section~\ref{sec2-evaluation}. Besides, we recruit eight qualified college students and ask them to solve the problems in \dataset independently, following the same pipeline of \engine. This score serves as a baseline for human performance. We conduct all experiments on NVIDIA A100 GPUs.


The input image dimensions for the webpage's top section screenshot are set to $1024 \times 1024$ pixels. For the full-page screenshot, we set the initial webpage width to $512$ pixels, although the actual width of a small portion of webpages may vary due to its layout settings. Furthermore, considering that a full-page screenshot can be extremely lengthy, directly inputting it as a single image into an LLM would result in excessive downsizing, making the content too vague for accurate identification. To address this, we segment the full-page screenshot into multiple images, starting from the top, with each segment measuring $512$ pixels in height. Because of the context length limitations of LMMs, the maximum number of full-page screenshot segments is therefore restricted to ten.

For the default settings, the longest edge of the input image is resized to match the largest resolution of the vision encoder of LMM. This ensures the image not to be cropped to multiple images and will only take up the minimum of tokens for image input. For any resolution settings, we input the image without resizing.

\begin{table*}[!t]
\small
\centering
\caption{\textbf{Evaluation Results on Different Subfields in \dataset.} SPO: Traditional Sports; ESP: E-Sports; ENT: Entertainment; GEN: General News; PAP: Paper; TEC: Technology; FIN: Finance; FAL: False Premise; ART: Arts; ARC: Architecture; AST: Astronomy; ANI: Anime; AUT: Auto; FAS: Fashion. The highest scores for \colorbox{backred!50}{closed-source} and \colorbox{backblue!75}{open-source} LMMs are marked in red and blue. \textit{For open-source LMMs, we adopt the models with 7B sizes unless otherwise specified.}}
\begin{adjustbox}{width=\linewidth}
 \renewcommand\tabcolsep{2.5pt} 
 \renewcommand\arraystretch{1.25} 
    \begin{tabular}{p{5.4cm}|>{\centering\arraybackslash}p{1cm}|c|cccccccc|c|cccccc}
    \toprule
    \multirow{3}*{\makecell*[l]{\large Model}}    
    &\multirow{3}*{\makecell*[c]
    {All}}
    &\multicolumn{9}{c|}{\makecell*[c]{\shortstack{News}}} 
    &\multicolumn{7}{c}{\makecell*[c]{\shortstack{Knowledge}}}\\
    \cmidrule{3-18}
    & &\makecell*[c]{Avg} &\makecell*[c]{SPO} &\makecell*[c]{ESP} &\makecell*[c]{ENT} &\makecell*[c]{GEN} &\makecell*[c]{PAP} &\makecell*[c]{TEC} &\makecell*[c]{FIN} &\makecell*[c]{FAL} &\makecell*[c]{Avg} &\makecell*[c]{ART}  &\makecell*[c]{ARC} &\makecell*[c]{AST} &\makecell*[c]{ANI} &\makecell*[c]{AUT} &\makecell*[c]{FAS} \\
    \midrule
    \multicolumn{18}{c}{\textit{Closed-source LMMs~~ with \engine}}\\
    \midrule
    Claude 3.5 Sonnet~\citep{anthropic2024claude35} & \cellcolor{verylightgray}53.5 & \cellcolor{verylightgray}53.0 & 37.4 & 50.2 & 63.6 & 49.2 & \cellcolor{backred!50}52.8 & 67.7 & 43.4 & 63.1 & \cellcolor{verylightgray}54.7 & 59.0 & 42.9 & \cellcolor{backred!50}70.9 & 56.5 & 60.6 & 36.4 \\
    GPT-4V~\citep{openai2023gpt4v} & \cellcolor{verylightgray}55.0 & \cellcolor{verylightgray}54.9 & 49.3 & 48.5 & 67.4 & 43.5 & 37.3 & 64.2 & 59.1 & \cellcolor{backred!50}90.2 & \cellcolor{verylightgray}55.3 & 54.6 & 45.6 & 65.1 & 63.0 & 52.2 & 56.2 \\
    GPT-4o~\citep{openai2024gpt4o} & \cellcolor{verylightgray}\cellcolor{backred!50}62.3 & \cellcolor{backred!50}61.2 & \cellcolor{backred!50}63.7 & \cellcolor{backred!50}61.2 & \cellcolor{backred!50}72.3 & \cellcolor{backred!50}51.3 & 48.6 & \cellcolor{backred!50}68.6 & \cellcolor{backred!50}60.0 & 76.8 & \cellcolor{backred!50}65.3 & \cellcolor{backred!50}73.8 & \cellcolor{backred!50}52.0 & 57.6 & \cellcolor{backred!50}76.8 & \cellcolor{backred!50}68.4 & \cellcolor{backred!50}55.8 \\
    \midrule
    \multicolumn{18}{c}{\textit{Open-source LMMs~~ with \engine}}\\
    \midrule
    InternLM-XC2.5~\citep{zhang2024internlm} & \cellcolor{verylightgray}22.2 & \cellcolor{verylightgray}22.8 & 22.6 & 13.6 & 28.9 & 23.5 & 15.5 & 27.8 & 32.9 & 3.1 & \cellcolor{verylightgray}20.5 & 29.2 & 24.0 & \cellcolor{verylightgray}20.8 & 2.0 & 20.4 & 11.8 \\
    InternLM-XC2.5\textit{\textsubscript{\textit{AnyRes}}} & \cellcolor{verylightgray}22.2 & \cellcolor{verylightgray}23.9 & 25.2 & 12.8 & 30.6 & 21.0 & 18.7 & 31.4 & 32.8 & 13.9 & \cellcolor{verylightgray}17.6 & 22.9 & 25.6 & 10.8 & 2.0 & 28.1 & 4.7 \\
    Mantis~\citep{jiang2024mantis} & \cellcolor{verylightgray}18.8 & \cellcolor{verylightgray}19.8 & 17.4 & 9.7 & 25.3 & 19.6 & 22.1 & 35.3 & 18.2 & 6.2 & \cellcolor{verylightgray}15.9 & 23.0 & 17.5 & 12.5 & 14.8 & 12.7 & 3.3 \\
    LLaVA-NeXT-Interleave~\citep{li2024llava-inter} & \cellcolor{verylightgray}28.3 & \cellcolor{verylightgray}29.3 & 23.3 & 18.5 & 41.9 & 33.6 & 26.2 & 27.3 & 28.0 & 14.7 & \cellcolor{verylightgray}25.6 & 31.1 & 22.7 & 49.7 & 16.5 & 19.7 & 7.1 \\
    InternVL2~\citep{chen2024far} & \cellcolor{verylightgray}34.3 & \cellcolor{verylightgray}35.7 & 38.7 & 20.7 & 46.8 & 33.2 & 24.6 & 44.1 & 44.5 & 27.2 & \cellcolor{verylightgray}30.1 & 41.7 & {33.5} & 26.9 & 15.4 & 36.8 & 7.9 \\
    InternVL2\textsubscript{\textit{AnyRes}} & \cellcolor{verylightgray}34.0 & \cellcolor{verylightgray}34.2 & 32.0 & 26.6 & 45.4 & 35.0 & 18.7 & 23.7 & 43.1 & 36.3 & \cellcolor{verylightgray}33.7 & 49.5 & 26.4 & 36.5 & 18.3 & 35.2 & 15.5 \\
    mPlug-Owl3~\citep{ye2024mplug} & \cellcolor{verylightgray}32.1 & \cellcolor{verylightgray}34.8 & 30.0 & 20.6 & 46.7 & 33.7 & 22.3 & 30.4 & 45.7 & {41.5} & \cellcolor{verylightgray}24.4 & 28.7 & 24.6 & 33.5 & 18.3 & 24.2 & 10.5 \\
    mPlug-Owl3\textsubscript{\textit{AnyRes}} & \cellcolor{verylightgray}33.9 & \cellcolor{verylightgray}35.5 & 42.0 & 27.1 & 49.7 & 34.4 & 24.5 & 30.4 & 41.3 & 18.9 & \cellcolor{verylightgray}29.3 & 34.4 & 20.3 & 40.9 & 18.8 & 32.8 & {24.4} \\
    Idefics3~\citep{laurençon2024building} & \cellcolor{verylightgray}36.2 & \cellcolor{verylightgray}38.5 & 43.5 & 24.2 & 50.1 & 41.7 & 27.5 & 36.2 & 38.8 & 37.4 & \cellcolor{verylightgray}29.6 & 34.2 & 32.1 & 32.1 & 29.6 & 32.7 & 9.1 \\
    Idefics3\textsubscript{\textit{AnyRes}} & \cellcolor{verylightgray}35.7 & \cellcolor{verylightgray}38.2 & 40.5 & 27.3 & 48.0 & 41.4 & 33.1 & 36.9 & 42.1 & 17.4 & \cellcolor{verylightgray}28.8 & 40.5 & 26.2 & 27.1 & 14.2 & 40.8 & 7.3 \\
    LLaVA-OneVision~\citep{li2024llava-ov} & \cellcolor{verylightgray}36.6 & \cellcolor{verylightgray}39.4 & 31.8 & 27.1 & 50.0 & 38.5 & 31.4 & 52.1 & 47.4 & 23.5 & \cellcolor{verylightgray}28.9 & 38.0 & 26.4 & 34.3 & 21.0 & 29.6 & 11.2 \\
    Qwen2-VL\textsubscript{\textit{AnyRes}}~\citep{Qwen2-VL} & \cellcolor{verylightgray}45.3 & \cellcolor{verylightgray}44.1 & 47.8 & 33.9 & 49.4 & \cellcolor{backblue!75}45.8 & 45.6 & 38.1 & 49.8 & 31.1 & \cellcolor{verylightgray}48.7 & \cellcolor{backblue!75}70.0 & 32.7 & 43.5 & 42.5 & 54.4 & 23.3 \\
    LLaVA-OneVision~(72B) & \cellcolor{verylightgray}50.1 & \cellcolor{verylightgray}52.3 & \cellcolor{backblue!75}53.2 & 45.4 & 62.1 & 45.6 & 41.5 & \cellcolor{backblue!75}64.8 & 47.8 & 37.2 & \cellcolor{verylightgray}44.5 & 63.4 & 42.8 & 52.4 & 24.3 & \cellcolor{backblue!75}70.0 & 31.7 \\
    Qwen2-VL\textsubscript{\textit{AnyRes}}~(72B) & \cellcolor{backblue!75}52.7 & \cellcolor{backblue!75}52.9 & 50.0 & \cellcolor{backblue!75}46.3 & \cellcolor{backblue!75}66.2 & 36.3 & \cellcolor{backblue!75}55.0 & 57.2 & \cellcolor{backblue!75}56.6 & \cellcolor{backblue!75}52.0 & \cellcolor{backblue!75}52.1 & 58.0 & \cellcolor{backblue!75}45.8 & \cellcolor{backblue!75}65.4 & \cellcolor{backblue!75}45.7 & 68.6 & \cellcolor{backblue!75}41.9 \\
    \bottomrule
    \end{tabular}
\end{adjustbox}
\label{table:main_result_topic}
\end{table*}

\subsection{Experimental Analysis}
\label{sec3-experiment_analysis}
To thoroughly investigate the multimodal searching capabilities, we present the evaluation results of different models on \dataset~following the proposed step-wise evaluation strategy in Table~\ref{table:main_result} and fourteen subfields in Table~\ref{table:main_result_topic}. We now provide a detailed discussion of notable findings and their implications for multimodal search capabilities. 


\textbf{Any-resolution input only provides slight or no improvement.}
Among the tested models, InternLM-XC2.5, InternVL2, mPlug-Owl3, and Idefic3 support both low-resolution (LowRes) and any-resolution input (AnyRes).
As one would expect, AnyRes input enables better OCR and perception of the image. However, we only observe slight or even no enhancement comparing the difference between the LowRes performance and its AnyRes counterpart. Take mPlug-Owl3 as an example, AnyRes input surpasses LowRes input on overall score by 1.8\%, end-to-end score by 2.7\%, and rerank by 0.2\%. While it falls behind LowRes on requery by 0.8\% and summarization by 1.7\%. This suggests that the OCR and perception quality do not bottleneck the search performance. Rather, the suboptimal performance appears to stem from the LMMs' inherent lack of robust search capabilities.

\textbf{Current LMMs still have significant shortcomings in requery and rerank.}
Comparing the average score of the end-to-end task with that of the summarization task, we find that the summarization score consistently surpasses the end-to-end task by a large margin, both in the closed-source and open-source models. The minimum margin is 2.7\% for GPT-4o, while the maximum is 23.9\% for LLaVA-OneVision-7B. 
This discrepancy can be attributed to the differences in the tasks' input quality. While the summarization task input always contains the answer, the end-to-end task's third-round input quality depends on the model's requery and rerank quality in previous rounds.
The magnitude of this performance gap reflects the disparity between a model's summarization ability and its capacity for requery and rerank tasks. The larger the difference, the larger the capability gap. Observing the result, we find that this gap of most open-sourced models exceeds 14\%, while the closed-sourced models are all below 10\%. This suggests all current LMMs need improvement of their requery and rerank ability, especially for open-source models. Mantis is one exception of open-source models with a margin of only 3.4\%. This means its poor summarization capability bottlenecks its end-to-end performance. Qwen2-VL-72B's 10.5\% gap, also falling below 14\%, highlights its superiority among other open-source LMMs.

\textbf{Closed-source LMMs are better-performed than open-sourced LMMs on overall performance.}
For the final score, closed-source LMMs consistently outperform the open-source ones. GPT-4o achieves the highest overall score of 62.3\%, demonstrating superior zero-shot multimodal search capabilities. While Qwen2-VL-72B leads among open-source models, it still lags behind GPT-4o by 9.6\%. The performance gap widens to 11.3\% on the most challenging end-to-end task and further expands to 20.1\% for 7B open-source LMMs. These significant disparities highlight substantial room for improvement in open-source models.

\textbf{SoTA LMMs with our \engine~surpass commercial AI search engines in the end-to-end task.}
We also evaluate the pro version of Perplexity~\citep{perplexity}, a prominent commercial AI search engine that accepts both image and text queries, on our dataset. Perplexity pro can accept both image and text in the user query. Surprisingly, although Perplexity also leverages SoTA LMMs like GPT-4o and Claude 3.5 Sonnet, it largely underperforms \engine~equipped with the same model in the end-to-end task. Even more remarkably, \engine can even surpass Perplexity with Qwen2-VL-72B, an open-source LMM. \textbf{\textit{This suggests that our \engine provides a better open-source plan for the multimodal AI search engine.}} The performance gap validates \engine's design effectiveness and highlights the value of testing various LMMs within our pipeline, since the pipeline can indeed achieve remarkable performance when using powerful LMMs. Upon investigating Perplexity's sub-optimal performance, we discover that it appears to utilize only a rudimentary image search algorithm, if any. This limitation leads to poor identification of the key objects in the image and failure to retrieve relevant information. Our findings underscore the effectiveness of \engine's design, particularly the incorporation of a robust image search step, which plays a crucial role in accurately recognizing important information from the input image.

\subsection{Error Analysis}
\label{sec3-error_analysis}
To investigate the limitations of current LMM search capabilities, we conduct a comprehensive analysis of error types observed in our evaluation. Our proposed step-wise evaluation strategy enables analysis of failure modes for each core search step, complementing the end-to-end assessment. This analysis encompasses the entire benchmark. We first examine the end-to-end error types for both the best-performing closed-source model (GPT-4o) and open-source model (Qwen2-VL-7B). To better understand the failure cases, we then identify distinct error types in the requery and summarization task, which requires open-ended generation. We quantify these error types for a systematic understanding of current LMM limitations and point out critical areas for improvement.
\vspace{0.2cm}


\begin{figure*}[ht]
\centering
\begin{minipage}[c]{0.46\textwidth}
\centering
\vspace{-0.1cm}
\includegraphics[width=0.82\textwidth]{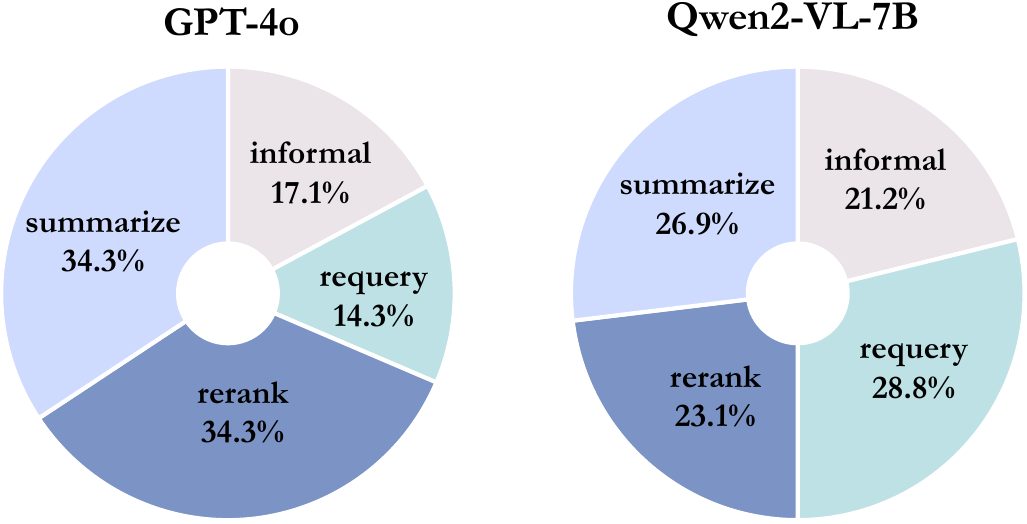} 
\caption{{Distribution of Error types of GPT-4o~\citep{openai2024gpt4o} and Qwen2-VL-7B~\citep{Qwen2-VL} in the end-to-end task.}}
\label{fig:error_analysis_e2e}
\end{minipage}
\qquad
\begin{minipage}[c]{0.46\textwidth}
\centering
\vspace{-0.2cm}
\label{fig:error_analysis_reqsum}
\vspace{0.15cm}
\includegraphics[width=\linewidth]{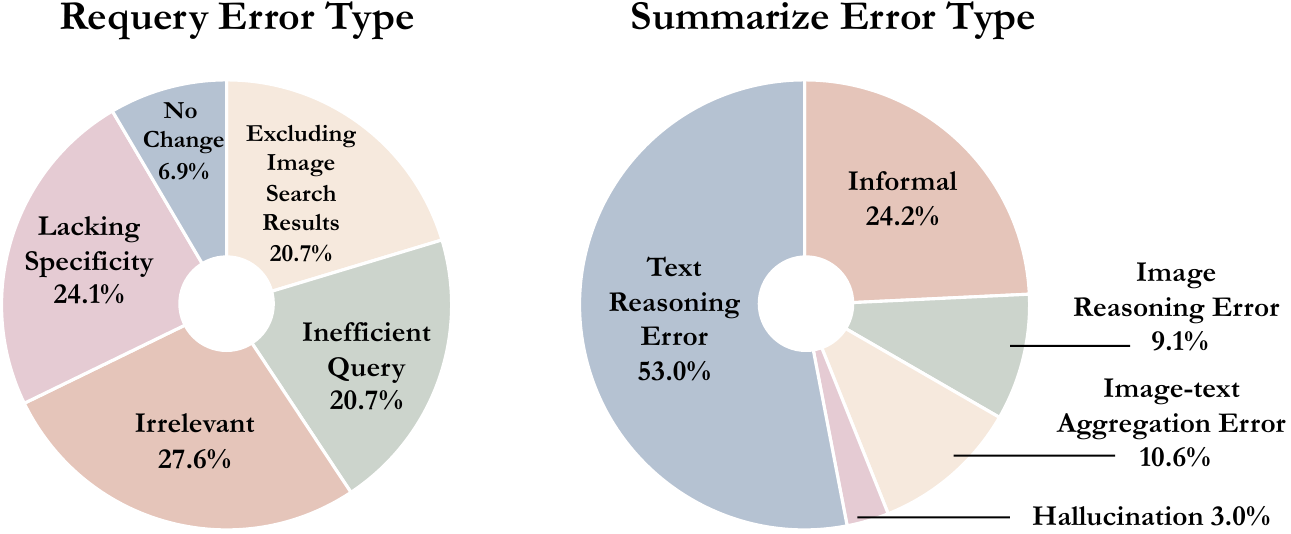}
\caption{{Distribution of Error types of Qwen2-VL-7B~\citep{Qwen2-VL} in the reuqery and summarization task respectively.}}
\end{minipage}
\label{fig:error_analysis}
\end{figure*}

\subsubsection{Error Analysis of End-to-end Task}
In this section, we are trying to answer the question: \textit{Which step does LMM make a mistake in the end-to-end evaluation?}~
In Fig.~\ref{fig:error_analysis_e2e}, we showcase the statistics of different error types occurring in GPT-4o and Qwen2-VL-7B. We define the following four error categories: (i)~\textit{requery}, where the model requery is incorrect, and leads to all retrieved websites being invalid; (ii)~\textit{rerank}, where the model selects a website without a correct answer; (iii)~\textit{summarization}, where the full website content contains the information of correct answer, but the model fails to extract it; (iv)~\textit{informal}, the output format deviates from the prompt specifications. 
As shown in the figure, GPT-4o's primary error sources are rerank and summarization errors, while requery and informal errors account for approximately half the frequency of the main error causes. This suggests that GPT-4o's limitations lie primarily in information source ranking and multimodal information integration. As for Qwen2-VL, all four error types occur with similar frequency. The rise of the informal error portion may be attributed to the model's inferior instruction-following ability. Besides, it should be noted that the requery task demands advanced comprehension and key image information extraction ability. This task seldom appears in the training data of current LMMs. The prevalence of this error type in Qwen2-VL may indicate that it fails to generalize to adequately address this complex task. 

\subsubsection{Error Analysis of Reuqery and Summarization Task}
To better understand how open-source LMM makes the mistake, we dive into the requery and summarization tasks to find out the error patterns of Qwen2-VL-7B. We particularly select the two tasks requiring open-ended generation, which provides more information to identify the error. 

As for the requery task, we categorize five types of errors: 
\begin{itemize}
    \item \textit{Lacking Specifility}, where the model fails to include all the specific information in the requery and therefore leads to sub-optimal search results. For example, the query is asking the release date of Vision Pro in China. However, the model omits the condition of China and directly asks about the release date of Vision Pro. 
    \item \textit{Inefficient Query}: The model generates requeries without considering practical search scenarios, leading to ineffective search results. For example, given a query comparing whether both Van Gogh's Sunflowers and Antoni Clavé's Grand Collage are oil paintings, an efficient approach would focus on finding information about the lesser-known Antoni Clavé's Grand Collage, since Van Gogh's Sunflowers is commonly known to be an oil painting. Instead, the model inefficiently repeats the original comparative query, which is unlikely to match existing content or yield useful search results.
    \item \textit{Excluding Image Search Results}, where the model totally ignores the information in the screenshot of the image search results and therefore lacks important specific information in the requery. For example, the query is `When did this football player obtain the gold medal?' and provides an image of the player. The model is supposed to find out the player's name by viewing the image search result and raise a requery like `[PLAYER NAME] obtained the gold medal time'. However, the model fails to incorporate the player's name in the requery and definitely the retrieved websites will not include any helpful information.
    \item \textit{No Change}, where the model just uses the question as the query input to the search engine.
    \item \textit{Irrelevant}, where the model either matches wrong information from the image search result or mistakenly understands the query and outputs an irrelevant requery.
\end{itemize}
    
These error types of requery suggest that LMM often fails to fully understand the requery task and fails to aggregate all available information. Besides, the error type of inefficient query indicates that LMM has no clue of the real working scenario and query principles of search engines. 

As for the summarization task, we also identify five types of errors: 
\begin{itemize}
    \item \textit{Text Reasoning Error}, where the model fails to extract the answer from the website textual information.
    \item \textit{Image-text Aggregation Error}, where obtaining the answer needs combining the information from both images and texts. The model fails to do so.
    \item \textit{Image reasoning Error}, where the model fails to extract the answer from the image, and the answer can only be obtained from the image.
    \item \textit{Hallucination}~\citep{huang2023survey}, where the model provides an unfaithful answer that cannot be grounded in the given content. 
    \item \textit{Informal}, the output format does not follow the prompt specifications, the same error type in the end-to-end task.
\end{itemize}
The occurrence of the five types of summarization errors reflects that current LMMs still cannot correctly extract the given multimodal information to answer the query. The ability of content understanding still requires further enhancement for current LMMs.

\subsection{Scaling Test-Time Compute vs Scaling Model Size}
\label{sec3-ttc}
Recent works such as OpenAI o1~\citep{gpt4o1} have highlighted the critical role of scaling test-time computation in enhancing model performance. Our end-to-end task, which requires multiple Internet interactions, presents an opportunity to investigate the potential of scaling test-time computation compared to scaling model size.
To explore this, we conduct experiments using LLaVA-OneVision-7B~\citep{li2024llava-ov}, focusing on scaling test-time computation, and compare against LLaVA-OneVision-72B scaling in model size, which aims to provide insights into the relative benefits of increased inference computation versus increased model parameters.

\begin{wraptable}{r}{0.4\textwidth}
  \centering
  \small
    \caption{\textbf{Scaling Test-Time Compute vs Scaling Model Size.} `TTC' and $\mathbf{S}_{e2e}$ denote Test-Time Computation and the score of end-to-end task.} 
    \resizebox{\linewidth}{!}{
        \begin{tabular}{lcc}
         \toprule
         \textbf{Model} & \textbf{Inference Cost} & \textbf{ $\mathbf{S}_{e2e}$} \\
         \midrule
         LLaVA-OV-7B & 1 & 29.6 \\
         LLaVA-OV-7B (TTC) & $\thicksim$25 & 55.2 \\
         LLaVA-OV-72B & $\thicksim$6 & 44.9 \\
         \bottomrule
         \end{tabular}
     }
     \vspace{-0.1cm}
    \label{experiment:ttc}
\end{wraptable}

For scaling up the test-time computation, we adopt a multi-modal search strategy similar to best-of-N solution, where `N' denotes 25 in our settings.
Specifically, for LLaVA-OneVision-7B, we first prompt the model to generate a requery 5 times, from which we select the one with the highest requery score $\mathbf{S}_{req}$. This requery is then used to retrieve brief results from 8 websites from a search engine. The model is again prompted 5 times to select the most informative website. After removing duplicates from the selected websites, we extract the full website content from the remaining ones and prompt the model to answer 5 times, obtaining 25 end-to-end outputs in total. We compute the F1 score for each answer against the ground truth and take the maximum as the model's end-to-end score for the query. Table~\ref{experiment:ttc} shows that LLaVA-OneVision-7B (TTC) achieves the score of 55.2\% in the end-to-end task, significantly enhancing the original score of 29.6\%, which surpasses LLaVA-OneVision-72B's 44.9\% and GPT-4V's 52.1\%. This result reveals the substantial potential of scaling test-time computation, validating the effectiveness of this technique as introduced by OpenAI o1. Our findings provide valuable insights for future research in this domain, suggesting that increased inference computation may offer comparable or superior performance improvements to increased model size not only in math and code tasks, but also in multimodal search tasks.

\section{Conclusion}
In this paper, we investigate the potential of LMMs as multimodal AI search engines. We first design \engine, a streamlined pipeline, enabling zero-shot LMMs to perform multimodal searches. To comprehensively assess the search capabilities, we introduce \dataset, a benchmark comprising 300 queries across 14 subfields. Our evaluation methodology analyzes LMM search abilities step-by-step, facilitating a deeper understanding of their limitations.
Using \engine, we evaluate various closed-source and open-source LMMs, revealing that current models still fall short of human-level search proficiency. Through thorough error analysis, we identify specific patterns of failure in key search process steps, providing valuable insights for future improvements in LMM search ability.

\bibliography{iclr2025_conference}
\bibliographystyle{iclr2025_conference}

\newpage
\appendix
\section*{Appendix Overview}
\begin{itemize}
    \item Section~\ref{appendix:related}: Related work.
    \item Section~\ref{appendix:exp_details}: Additional experimental details.
    \item Section~\ref{appendix:data_details}: More dataset details.
    \item Section~\ref{appendix:qualitative}: Qualitative examples.
\end{itemize}

\vspace{0.3cm}
\section{Related work}
\label{appendix:related}
\paragraph{\textbf{{Large Multimodal Models.}}} 
Recently, multimodal models~\citep{Radford2021LearningTV, li2022blip, openai2023gpt4v, rombach2022high, jiang2024comat} has gained unparalleled attention. Building on the success of Large Language Models (LLMs)~\citep{touvron2023llama,touvron2023llama2} and large-scale vision models~\citep{Radford2021LearningTV}, Large Multimodal Models (LMMs) are gaining prominence across diverse domains. These models extend LLMs to handle tasks involving various modalities, including mainstream 2D image processing~\citep{liu2023llava,zhu2023minigpt,lin2023sphinx,gao2023llamaadapterv2}, as well as 3D point clouds~\citep{xu2023pointllm, guo2023point,guo2024sam2point}, and videos~\citep{li2023videochat, chen2023videollm, zhang2023video,fu2024video}. 
Among these LMMs, OpenAI's GPT-4o~\citep{openai2024gpt4o} and Anthropic's Claude 3.5 Sonnet~\citep{anthropic2024claude35} demonstrate outstanding visual reasoning and comprehension capability, setting new standards in multi-modal performance. However, their closed-source nature limits broader adoption and development. 
In contrast, another research trajectory focuses on open-source LMMs for the community. Pioneering works like LLaVA~\citep{liu2023llava,liu2024llavanext,li2024llava-inter,li2024llava-ov}, LLaMA-Adapter~\citep{zhang2024llamaadapter, gao2023llamaadapterv2}, and MiniGPT-4~\citep{zhu2023minigpt, chen2023minigpt} incorporate a frozen CLIP~\citep{Radford2021LearningTV} model for image encoding and integrate visual information into LLM for multi-modal instruction tuning. Later, works such as mPLUG-Owl~\citep{ye2023mplugowl, ye2023mplugowl2, ye2024mplug}, SPHINX~\citep{gao2024sphinx, lin2023sphinx}, and InternLM-XComposer~\citep{dong2024internlm} further advanced the field by incorporating diverse visual instruction tuning data and generalizing to more scenarios. More recent developments in the field have taken diverse directions. For example, several studies \citep{zong2024mova, tong2024eyes} explore multiple vision encoders design. Meanwhile, other works~\citep{liu2024llavanext, chen2024far, Qwen2-VL} incorporate high-resolution image input. Multi-image instruction data~\citep{li2024llava-inter, jiang2024mantis} is also integrated to enable perception across multiple images. 
While various benchmarks, both in the general~\citep{fu2023mme, liu2023mmbench, Yu2023MMVetEL} and expert~\citep{zhang2024mathverse, Lu2023MathVistaEM, lu2022learn} domain, has been proposed, the potential of LMM to function as a multimodal search engine remains largely unexplored. To this end, we introduce the \dataset benchmark, which evaluates LMMs' zero-shot abilities of multimodal search, offering valuable insights for future research.

\paragraph{\textbf{{Large models with Retrieval Augmented Generation (RAG).}}}
RAG (Retrieval-Augmented Generation) is an effective strategy for enhancing model knowledge by retrieving relevant information from external sources~\citep{fan2024survey}. RAG has been leveraged in various scenarios including knowledge-intensive question answering~\citep{borgeaud2022improving, guu2020retrieval}, machine translation~\citep{he2021fast}, and hallucination elimination~\citep{bechard2024reducing}. Current works has focused on improving specific aspects of RAG. RG-RAG~\citep{chan2024rq} proposes to refine the query for retrieval by decomposition and disambiguation. Self-RAG~\citep{asai2023self} incorporates the self-reflection of LLM to enhance the generation quality. The AI search engine could be viewed as a form of RAG with the Internet serving as the external knowledge source. Recently, MindSearch~\citep{chen2024mindsearch} proposes an AI search engine framework to simulate the human minds in web information seeking. Meanwhile, multiple benchmarks of RAG~\citep{yang2024crag, chen2024benchmarking} have been introduced to comprehensively evaluate a RAG system. However, both the current AI search engine and RAG benchmark are limited to the text-only setting, leaving the multimodal search engine and evaluation largely unexplored. To bridge this gap, we introduce \engine and \dataset, a multimodal AI search engine pipeline and dataset designed to evaluate various multimodal scenarios.

\clearpage
\section{Additional experimental details}
\label{appendix:exp_details}
\textbf{Model Sources.} For different LMMs, we select their latest models with size around 7B for evaluation to fully reveal their multimodal search proficiency. Table~\ref{table:release_time} presents the release time and model sources of LMMs used in \dataset.

\begin{table*}[h]
    \centering
    \caption{\textbf{The Release Time and Model Source of LMMs Used in \dataset.}}
    \resizebox{0.9\linewidth}{!}{ 
    \begin{tabular}{l@{\hspace{1cm}}cp{0.5\textwidth}}
    
    \toprule
    \textbf{Model} & \textbf{\makecell{Release\\ Time}} & \textbf{\makecell[c]{Source}} \\
    \midrule
    GPT-4V~\citep{openai2023gpt4v}     &   2023-09    & \url{https://platform.openai.com/} \\
    \midrule
    GPT-4o~\citep{openai2024gpt4o}     &   2024-05    & \url{https://platform.openai.com/} \\
    \midrule
    \multirow{2}{*}{Claude 3.5 Sonnet~\citep{anthropic2024claude35}}        & \multirow{2}{*}{2024-06}  & \url{https://www.anthropic.com/news/claude-3-5-sonnet} \\
    \midrule
    \multirow{2}{*}{InternLM-XC2.5~\citep{zhang2024internlm}}    & \multirow{2}{*}{2024-07}    & \url{https://github.com/InternLM/InternLM-XComposer} \\
    \midrule
    \multirow{2}{*}{Mantis~\citep{jiang2024mantis}}         & \multirow{2}{*}{2024-05} & \url{https://tiger-ai-lab.github.io/Mantis/} \\
     \midrule
    \multirow{2}{*}{LLaVA-NeXT-Interleave~\citep{li2024llava-inter}}      &   \multirow{2}{*}{2024-06}   & \url{https://github.com/LLaVA-VL/LLaVA-NeXT} \\
    \midrule
    \multirow{2}{*}{InternVL2~\citep{chen2024far}}       &   \multirow{2}{*}{2024-07}    &        \url{https://github.com/OpenGVLab/InternVL} \\
    \midrule
    \multirow{2}{*}{mPlug-Owl3~\citep{ye2024mplug}}         &    \multirow{2}{*}{2024-08}       & \url{https://github.com/X-PLUG/mPLUG-Owl} \\
    \midrule
    \multirow{2}{*}{Idefics3~\citep{laurençon2024building}}       &   \multirow{2}{*}{2024-08}    &        \url{https://huggingface.co/HuggingFaceM4/Idefics3-8B-Llama3} \\
    \midrule
    \multirow{2}{*}{LLaVA-OneVision~\citep{li2024llava-ov}}       &   \multirow{2}{*}{2024-08}    &        \url{https://llava-vl.github.io/blog/2024-08-05-llava-onevision/} \\
    \midrule
    \multirow{2}{*}{Qwen2-VL~\citep{Qwen2-VL}}       &   \multirow{2}{*}{2024-08}    &        \url{https://github.com/QwenLM/Qwen2-VL} \\
    \bottomrule
    \end{tabular}
    }
    \label{table:release_time}
\end{table*}

\begin{table*}[!ht]
    \centering
    \caption{\textbf{Input Prompt of LMMs for Requery.} We adopt two different prompts for the query with image input and without image input. We leverage a 3-shot prompt to guide the LMM to generate a reasonable requery.}
    \begin{tabular}{p{0.2\textwidth}p{0.7\textwidth}}
    \toprule
    \textbf{Question}                                       & \textbf{Prompt} \\
    \midrule
        \multirow{17}{*}{Query without image}  &You are a helpful assistant. I am giving you a question, which cannot be solved without external knowledge.
        Assume you have access to a text-only search engine (e.g., google). Please raise a query to the search engine to search for what is useful for you to answer the question correctly. Your query needs to consider the attribute of the query to search engine. 
        Here are 3 examples: \\
        & Question: Did Zheng Xiuwen wear a knee pad in the women's singles tennis final in 2024 Paris Olympics?
        Query to the search engine: Images of Zheng Xiuwen in the women's singles tennis final in 2024 Paris Olympics
        
        Question: When will Apple release iPhone16?
        Query to the search engine: iPhone 16 release date 
        
        Question: Who will sing a French song at the Olympic Games closing ceremony?
        Query to the search engine: Singers at the Olympic Games closing ceremony, French song. 
        
        Question: $\{question\}$. \\
        & Query to the search engine (do not involve any explanation): \\
    \midrule
        \multirow{21}{*}{Query with image}  &You are a helpful assistant. I am giving you a question including an image, which cannot be solved without external knowledge.
        Assume you have access to a search engine (e.g., google). Please raise a query to the search engine to search for what is useful for you to answer the question correctly. You need to consider the characteristics of asking questions to search engines when formulating your questions. 
        You are also provided with the search result of the image in the question. You should leverage the image search result to raise the text query.
        Here are 3 examples: \\
        & Question: Did Zheng Xiuwen wear a knee pad in the women's singles tennis final in 2024 Paris Olympics?
        Query to the search engine: Images of Zheng Xiuwen in the women's singles tennis final in 2024 Paris Olympics
        
        Question: When will Apple release iPhone16?
        Query to the search engine: iPhone 16 release date 
        
        Question: Who will sing a French song at the Olympic Games closing ceremony?
        Query to the search engine: Singers at the Olympic Games closing ceremony, French song \\

        & Question: $\{query\_image\}\{question\}$. The image search result is: $\{image\_search\_result\}$ \\
        & Query to the search engine (do not involve any explanation): \\
    \bottomrule
    \end{tabular}
    \label{appendix:prompt-requery}
\end{table*}

\begin{table*}[!ht]
    \centering
    \caption{\textbf{Input Prompt of LMMs for Rerank.} We adopt two different prompts for the query with image input and without image input.}
    \begin{tabular}{p{0.2\textwidth}p{0.7\textwidth}}
    \toprule
    \textbf{Question}                                       & \textbf{Prompt} \\
    \midrule
        \multirow{11}{*}{Query without image}  & You are a helpful assistant. I am giving you a question and 8 website information related to the question (including the screenshot, snippet and title). You should now read the screenshots, snippets and titles. Select 1 website that is the most helpful for you to answer the question. Once you select it, the detailed content of them will be provided to help you correctly answer the question. The question is $\{question\}$. The website informations is
        $\{website\_information\}$. \\
        & You should directly output 1 website's index that can help you most, and enclose the website in angle brackets. The output format should be: \textless Website Index \textgreater. 
        An example of the output is: \textless Website 1 \textgreater. 
        Your answer: \\
    \midrule
        \multirow{14}{*}{Query with image}  &You are a helpful assistant. I am giving you a question including an image. You are provided with the search result of the image in the question. And you are provided with 8 website information related to the question (including the screenshot, snippet, and title). 
        You should now read the screenshots, snippets and titles of these websites. Select 1 website that is the most helpful for you to answer the question. Once you select it, the detailed content of them will be provided to help you correctly answer the question. The question is $\{query\_image\}\{question\}$.
        The image search result is $\{image\_search\_result\}$.
        The website information is $\{website\_information\}$. \\
        & You should directly output 1 website's index that can help you most, and enclose the website in angle brackets. The output format should be: \textless Website Index \textgreater. 
        An example of the output is: \textless Website 1 \textgreater.
        Your answer: \\
    \bottomrule
    \end{tabular}
    \label{appendix:prompt-rerank}
\end{table*}

\begin{table*}[!ht]
    \centering
    \caption{\textbf{Input Prompt of LMMs for Summarization.} We adopt two different prompts for the query with image input and without image input.}
    \begin{tabular}{p{0.2\textwidth}p{0.7\textwidth}}
    \toprule
    \textbf{Question}                                       & \textbf{Prompt} \\
    \midrule
        \multirow{11}{*}{Query without image}  & You are a helpful assistant. I am giving you a question and 1 website information related to the question. 
        Please follow these guidelines when formulating your answer:
        1. If the question contains a false premise or assumption, answer ``invalid question''.
        2. When answering questions about dates, use the yyyy-mm-dd format.
        3. Answer the question with as few words as you can.
        
        You should now read the information of the website and answer the question.
        The website information is $\{website\_information\}$.
        The question is $\{question\}$.
        Please directly output the answer without any explanation: \\
    \midrule
        \multirow{14}{*}{Query with image}  &You are a helpful assistant. I am giving you a question including an image. You are provided with the search result of the image in the question. And you are provided with 1 website information related to the question. 
        Please follow these guidelines when formulating your answer:
        1. If the question contains a false premise or assumption, answer ``invalid question''.
        2. When answering questions about dates, use the yyyy-mm-dd format.
        3. Answer the question with as few words as you can.
        
        You should now read the information of the website and answer the question.
        The website information is $\{website\_information\}$.
        The image search result is $\{image\_search\_result\}$. 
        The question is $\{query\_image\}\{question\}$.
        Please directly output the answer without any explanation:  \\
    \bottomrule
    \end{tabular}
    \label{appendix:prompt-summarization}
\end{table*}

\textbf{Full-page Screenshot Slimming.}
For the full-page screenshot, we compute the Sobel gradients~\citep{kanopoulos1988design} to detect the edges and generate a gradient magnitude image. We iteratively remove the areas with gradients below a threshold, which represent the blank areas. This approach effectively reduces image size while maintaining critical document content. 

\begin{figure}[!h]
\includegraphics[width=\textwidth]{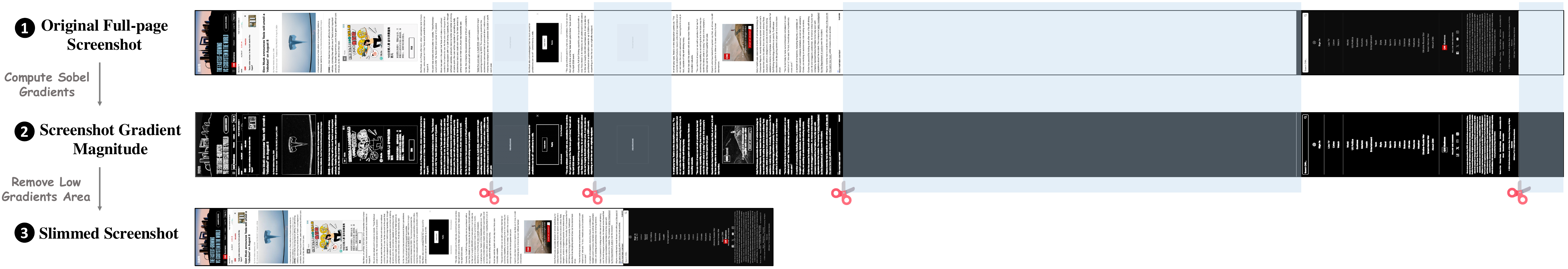} 
\caption{\textbf{Illustration of the Screenshot Slim Process.} We leverage Sobel gradients~\citep{kanopoulos1988design} to identify blank areas and remove them. After slimming, the screenshot size is largely reduced without any information loss.}
\label{fig:slim}
\end{figure}

\vspace{0.7cm}

\textbf{Input Prompts of LMM for Response Generation.}
We showcase the input prompts of LMM for the three tasks respectively in Table~\ref{appendix:prompt-requery}-\ref{appendix:prompt-summarization}. We adopt two types of prompts for queries with an image and without images. For query with an image, we specifically require the LMM to leverage the image search result to solve the task.

\clearpage
\section{More data details}
\label{appendix:data_details}
\subsection{Subfield definition}
\label{appendix:data_details_subfield}
\textbf{News}~area encompasses a vast spectrum of information, ranging from everyday events to engaging entertainment content and specialized fields such as scientific discoveries and financial analysis. This comprehensive coverage serves as a rigorous assessment of the model's ability to process information in diverse domains. We divide this expansive area into eight distinct subfields:

\begin{itemize}
    \item \textbf{Traditional Sports}:
    Data concerning traditional athletic competitions, team performances, player statistics, and sporting events. This includes scores, league standings, player transfers, and analysis of various professional sports across different leagues and countries.
    \item \textbf{e-Sports}:
    Information about competitive video gaming, including tournament results, player rankings, and league information. This covers various game titles, team formations, streaming viewership statistics, and tournament information.
    \item \textbf{Technology}:
    Information about technological innovations, gadgets, software developments, and tech industry news. This includes product launches, software updates, cybersecurity issues, and artificial intelligence advancements.
    \item \textbf{Paper}:
    Content related to academic papers, research publications, and scholarly articles in various artificial intelligence fields. The queries include method explanation, figure understanding, and experiment settings.
    \item \textbf{Entertainment}:
    Data about movies, music, television, celebrities, and other forms of popular entertainment. It also includes data concerning video games.
    \item \textbf{Finance}:
    Information on financial markets, economic indicators, business news, and monetary policies. This covers stock prices, company earnings reports, company financial statements, and regulatory news regarding finance.
    \item \textbf{General News}:
    Broad coverage of various news topics not specific to any particular subfield. This includes a mix of local and global events, human interest stories, lifestyle articles, climate news, and general interest content that doesn't fit neatly into other specialized news subfields.
    \item \textbf{False Premise}:
    Data related to misinformation or incorrect assumptions in the query. This subfield focuses on fact-checking capabilities. All the answers to the queries of this subfield are `invalid question'.
\end{itemize}

\textbf{Knowledge}~area represents broad subfields of information and data related to general knowledge across various disciplines. This area concentrates on rare knowledge that most LMMs fail to answer. We categorize this area into five subfields:

\begin{itemize}
    \item \textbf{Architecture}:
    Information about building design, architectural styles, building information, and construction projects. This includes city landmarks, the comparison of architectural styles, and multi-view architecture matchings.
    \item \textbf{Arts}:
    Data concerning visual arts, drawings, sculptures, badges, and other forms of creative expression. This covers artwork details, artist profiles, artwork history, and artwork style comparisons.
    \item \textbf{Fashion}:
    Content related to clothing trends, fashion brands, and designer collections. This includes retail price, clothing style, release date, and brand information.
    \item \textbf{Astronomy}:
    Information about celestial objects, space exploration, astronomical phenomena, and related research. This covers observational data from telescopes and image results from space missions. The questions focus on the background information of these celestial objects presented in the query image.
    \item \textbf{Anime}:
    Data about Japanese animation, including series storylines and character information. This encompasses character background, character appearance, voice actor information, and chapter information.
    \item \textbf{Auto}:
    Content related to automobiles, including vehicle specifications, industry trends, and automotive technology. This covers new car models, performance test results, coefficients of cars, and release date.
\end{itemize}

\clearpage
\section{Qualitative examples}
\label{appendix:qualitative}

\begin{figure}[!h]
\centering
\includegraphics[width=\textwidth]{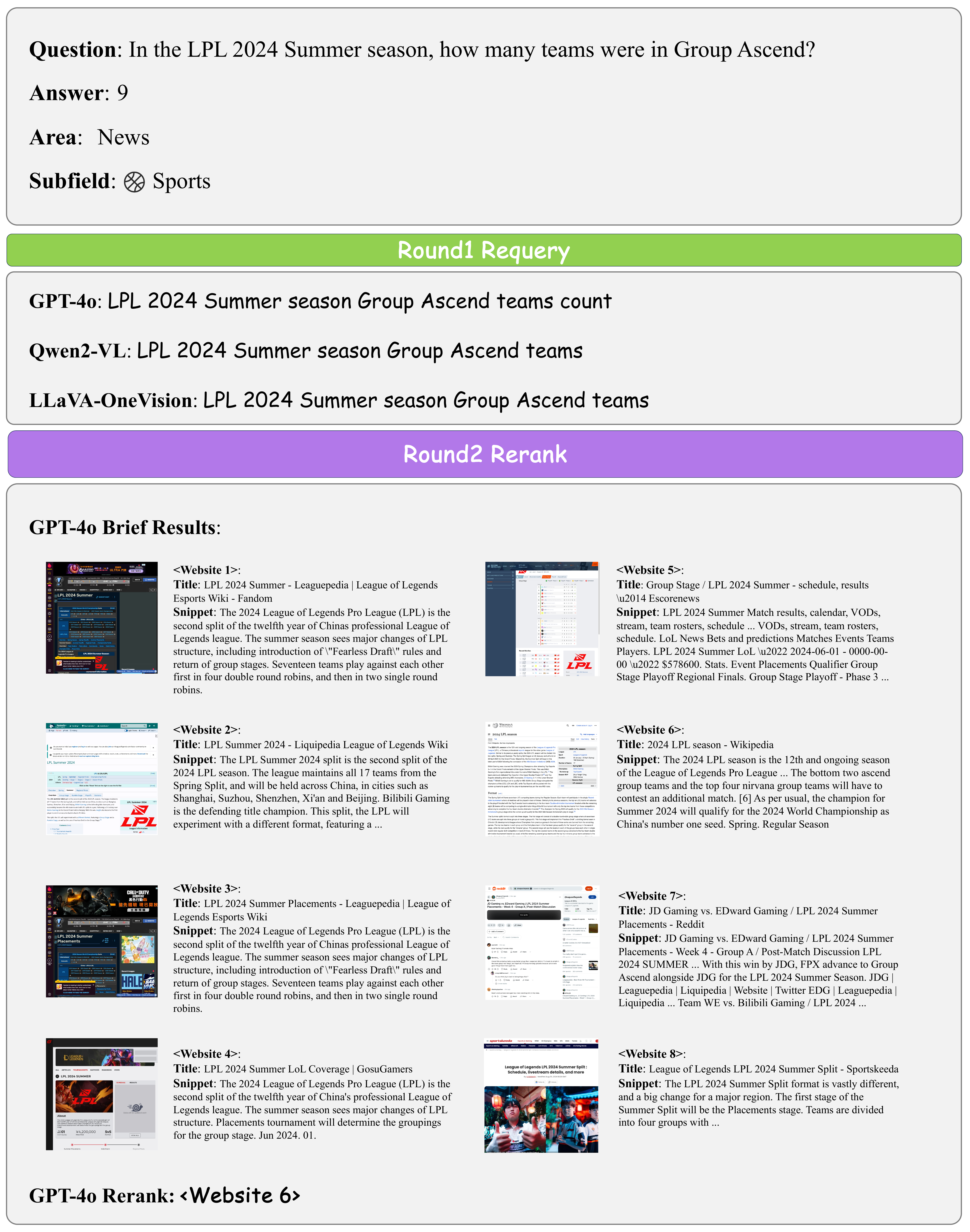} 
\caption{Response and middle results comparison of GPT-4o~\citep{openai2024gpt4o}, Qwen2-VL-7B~\citep{Qwen2-VL}, and LLaVA-OneVision-7B~\citep{li2024llava-ov} in the end-to-end task.}
\label{fig:data_sample_exp1_p1}
\end{figure}

\begin{figure}[ht]
\centering
\includegraphics[width=\textwidth]{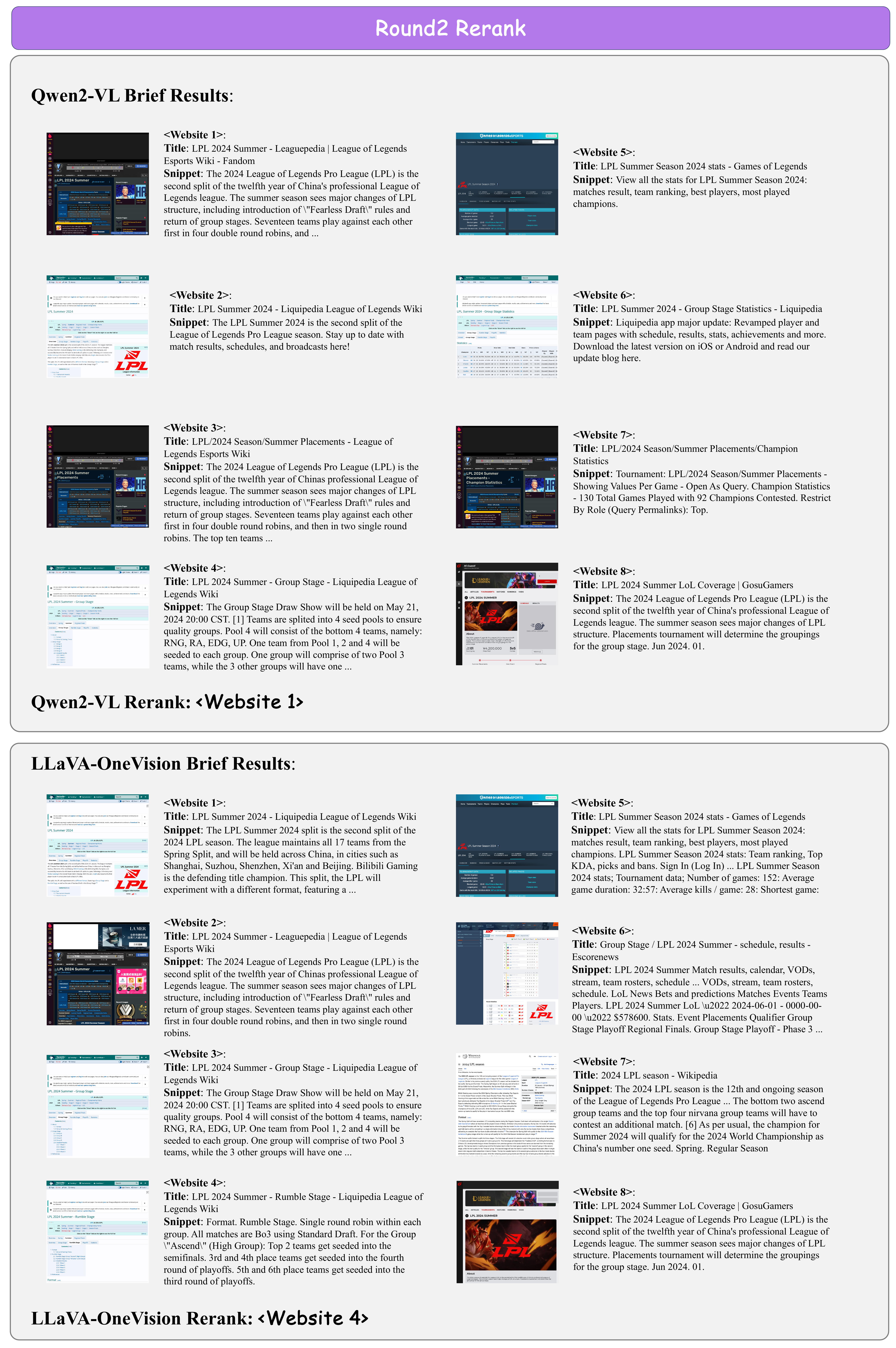} 
\caption{Response and middle results comparison of GPT-4o~\citep{openai2024gpt4o}, Qwen2-VL-7B~\citep{Qwen2-VL}, and LLaVA-OneVision-7B~\citep{li2024llava-ov} in the end-to-end task.}
\label{fig:data_sample_exp1_p1}
\end{figure}

\begin{figure}[ht]
\centering
\includegraphics[width=\textwidth]{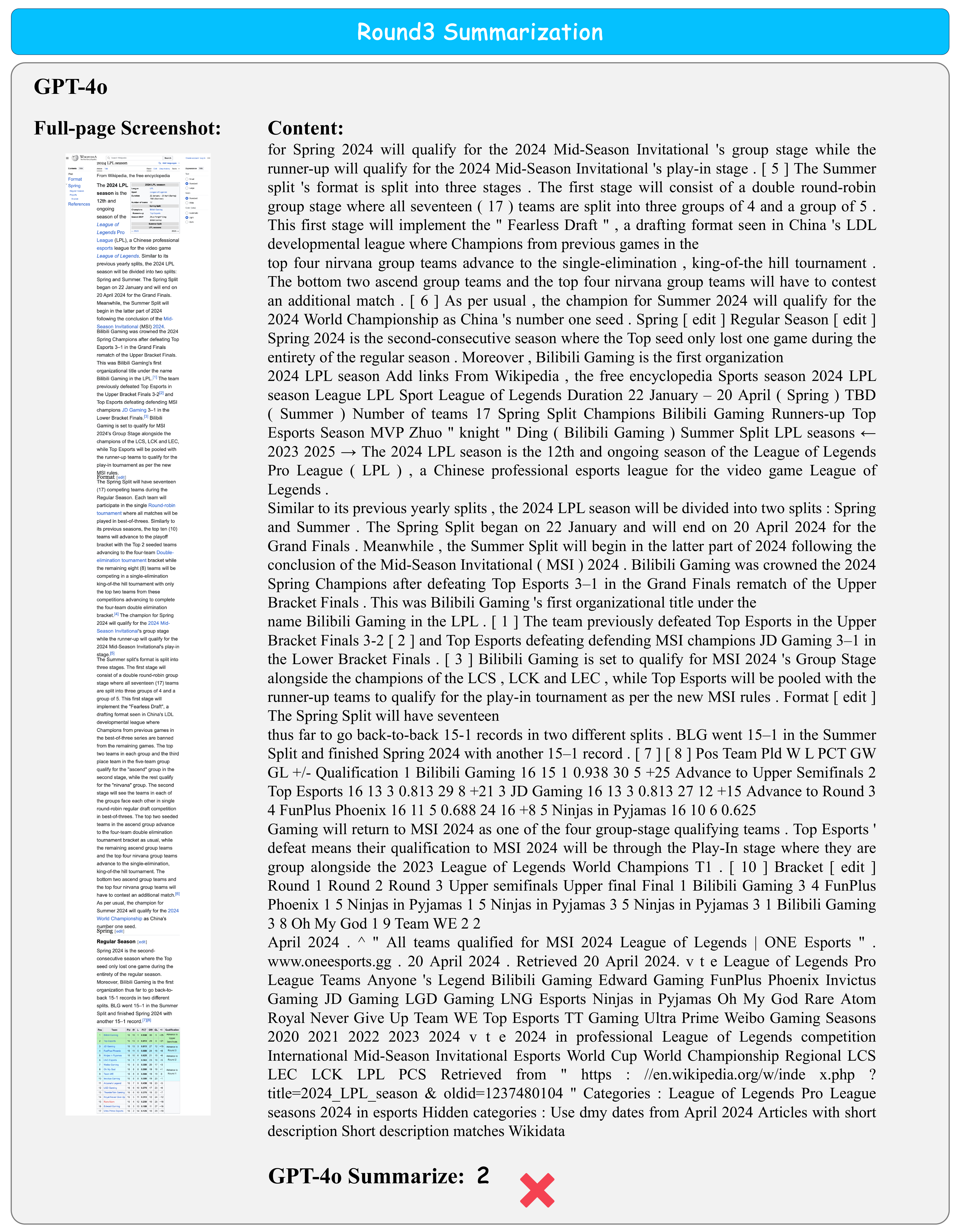} 
\caption{Response and middle results comparison of GPT-4o~\citep{openai2024gpt4o}, Qwen2-VL-7B~\citep{Qwen2-VL}, and LLaVA-OneVision-7B~\citep{li2024llava-ov} in the end-to-end task.}
\label{fig:data_sample_exp1_p1}
\end{figure}

\begin{figure}[ht]
\centering
\includegraphics[width=\textwidth]{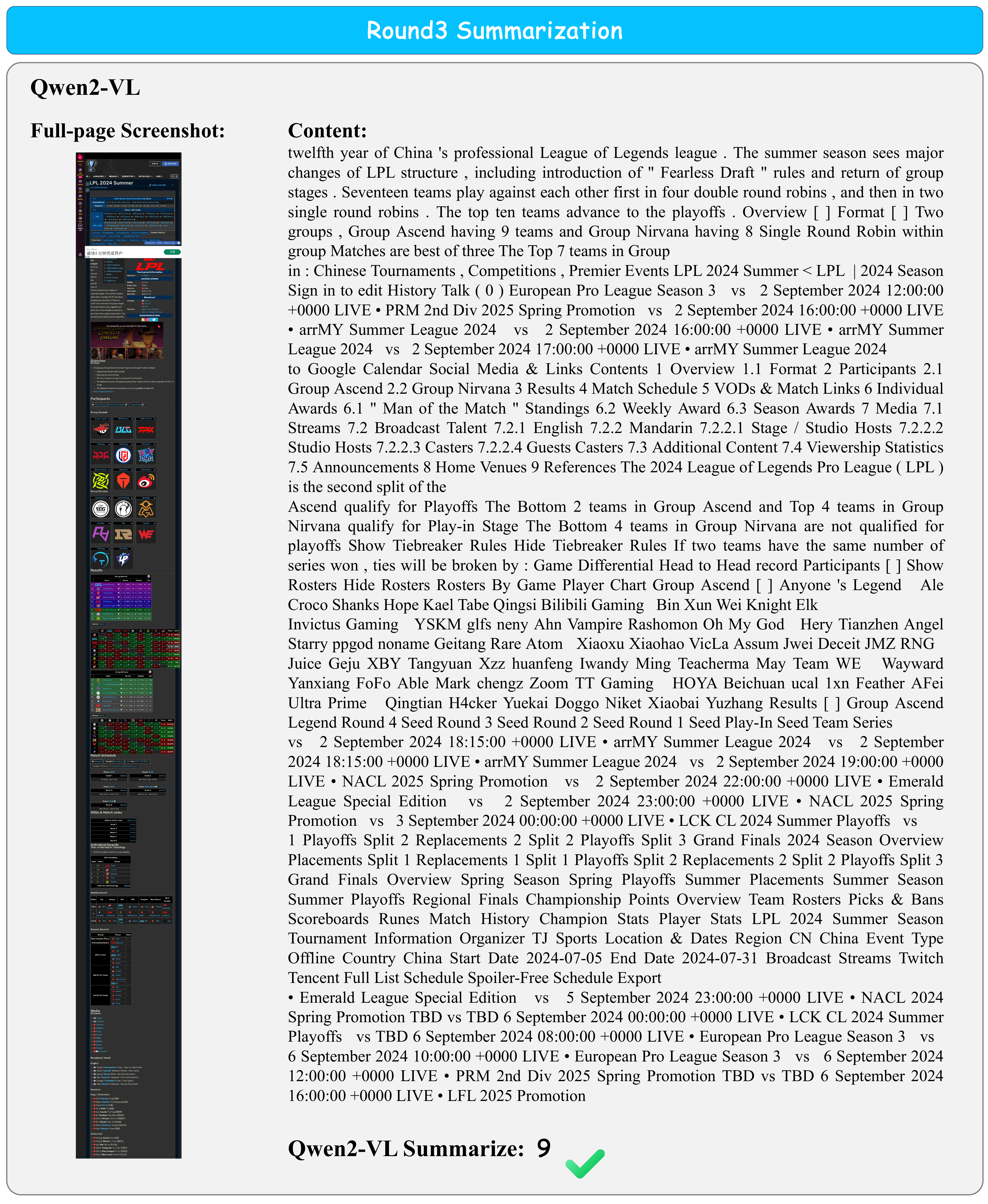} 
\caption{Response and middle results comparison of GPT-4o~\citep{openai2024gpt4o}, Qwen2-VL-7B~\citep{Qwen2-VL}, and LLaVA-OneVision-7B~\citep{li2024llava-ov} in the end-to-end task.}
\label{fig:data_sample_exp1_p1}
\end{figure}

\begin{figure}[ht]
\centering
\includegraphics[width=\textwidth]{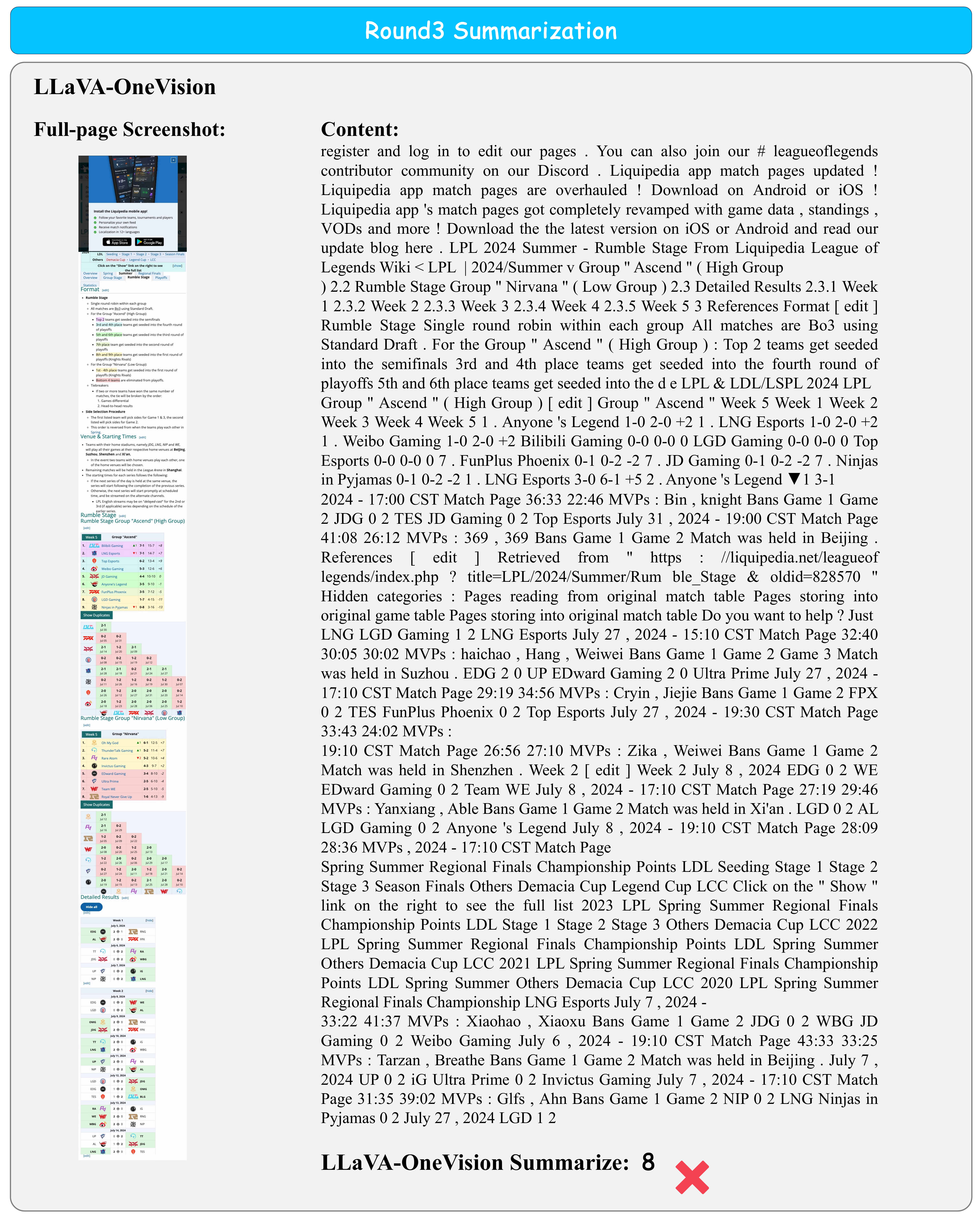} 
\caption{Response and middle results comparison of GPT-4o~\citep{openai2024gpt4o}, Qwen2-VL-7B~\citep{Qwen2-VL}, and LLaVA-OneVision-7B~\citep{li2024llava-ov} in the end-to-end task.}
\label{fig:data_sample_exp1_p1}
\end{figure}

\begin{figure}[ht]
\centering
\includegraphics[width=\textwidth]{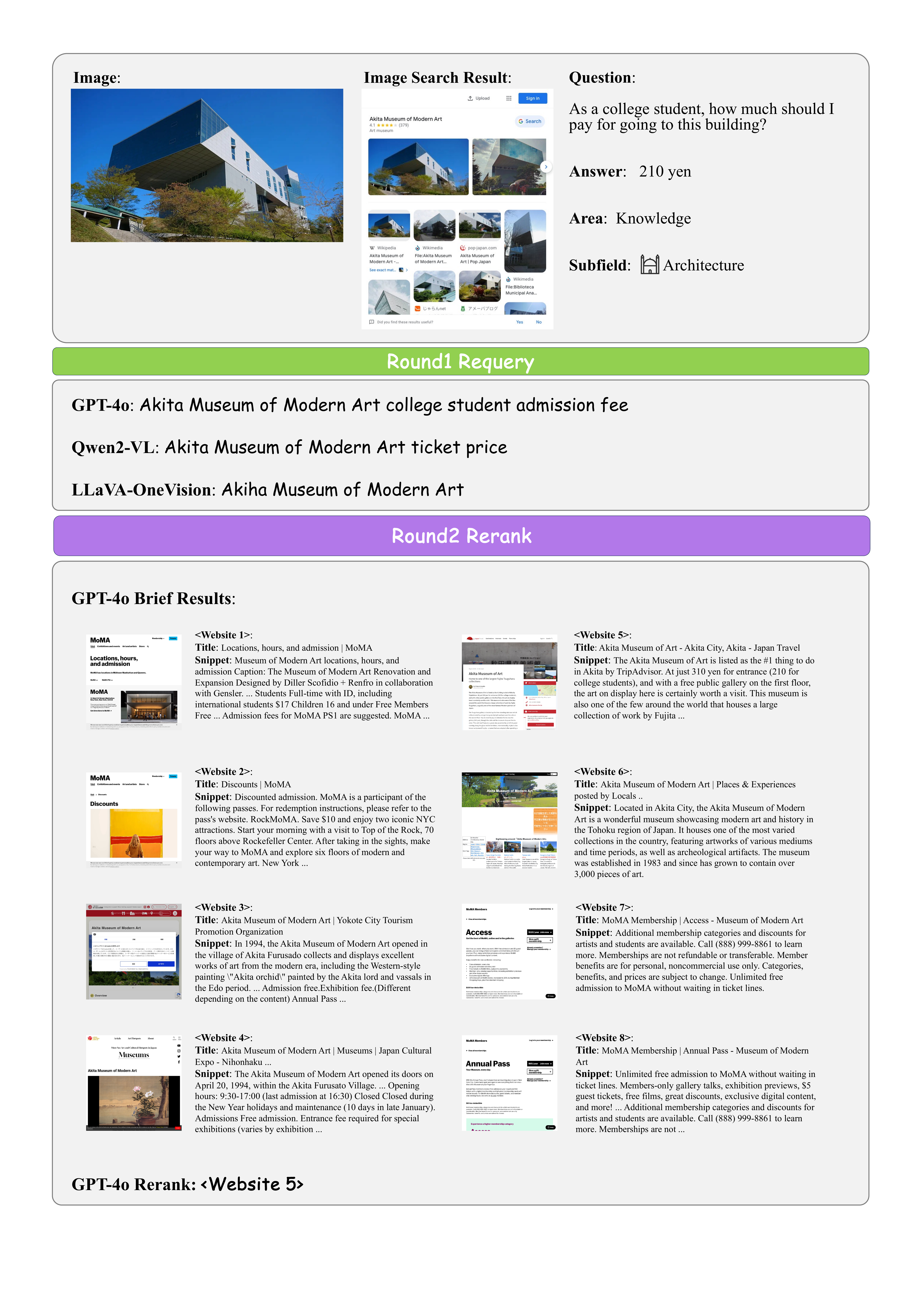} 
\caption{Response and middle results comparison of GPT-4o~\citep{openai2024gpt4o}, Qwen2-VL-7B~\citep{Qwen2-VL}, and LLaVA-OneVision-7B~\citep{li2024llava-ov} in the end-to-end task.}
\label{fig:data_sample_exp1_p1}
\end{figure}

\begin{figure}[ht]
\centering
\includegraphics[width=\textwidth]{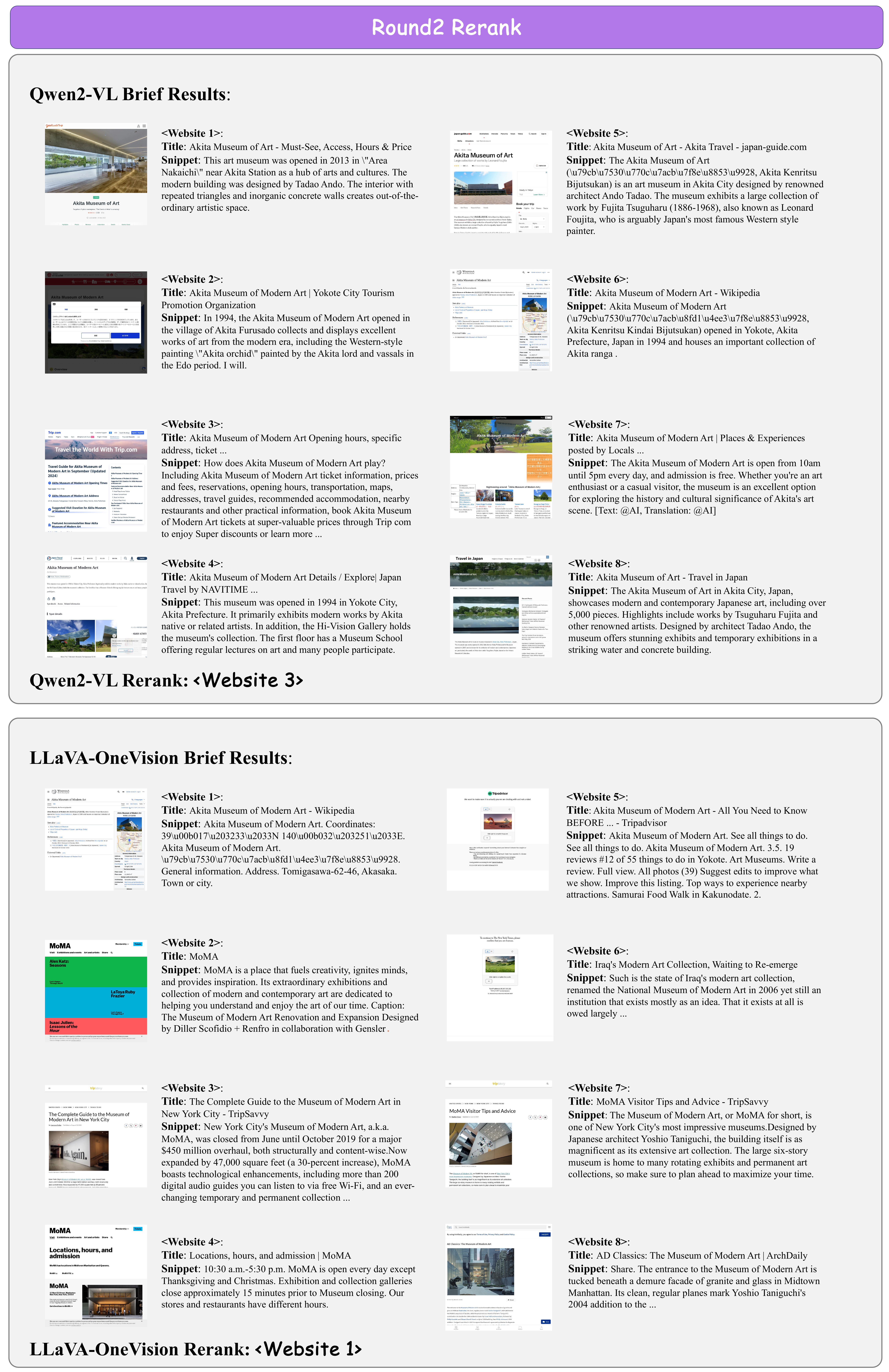} 
\caption{Response and middle results comparison of GPT-4o~\citep{openai2024gpt4o}, Qwen2-VL-7B~\citep{Qwen2-VL}, and LLaVA-OneVision-7B~\citep{li2024llava-ov} in the end-to-end task.}
\label{fig:data_sample_exp1_p1}
\end{figure}

\begin{figure}[ht]
\centering
\includegraphics[width=\textwidth]{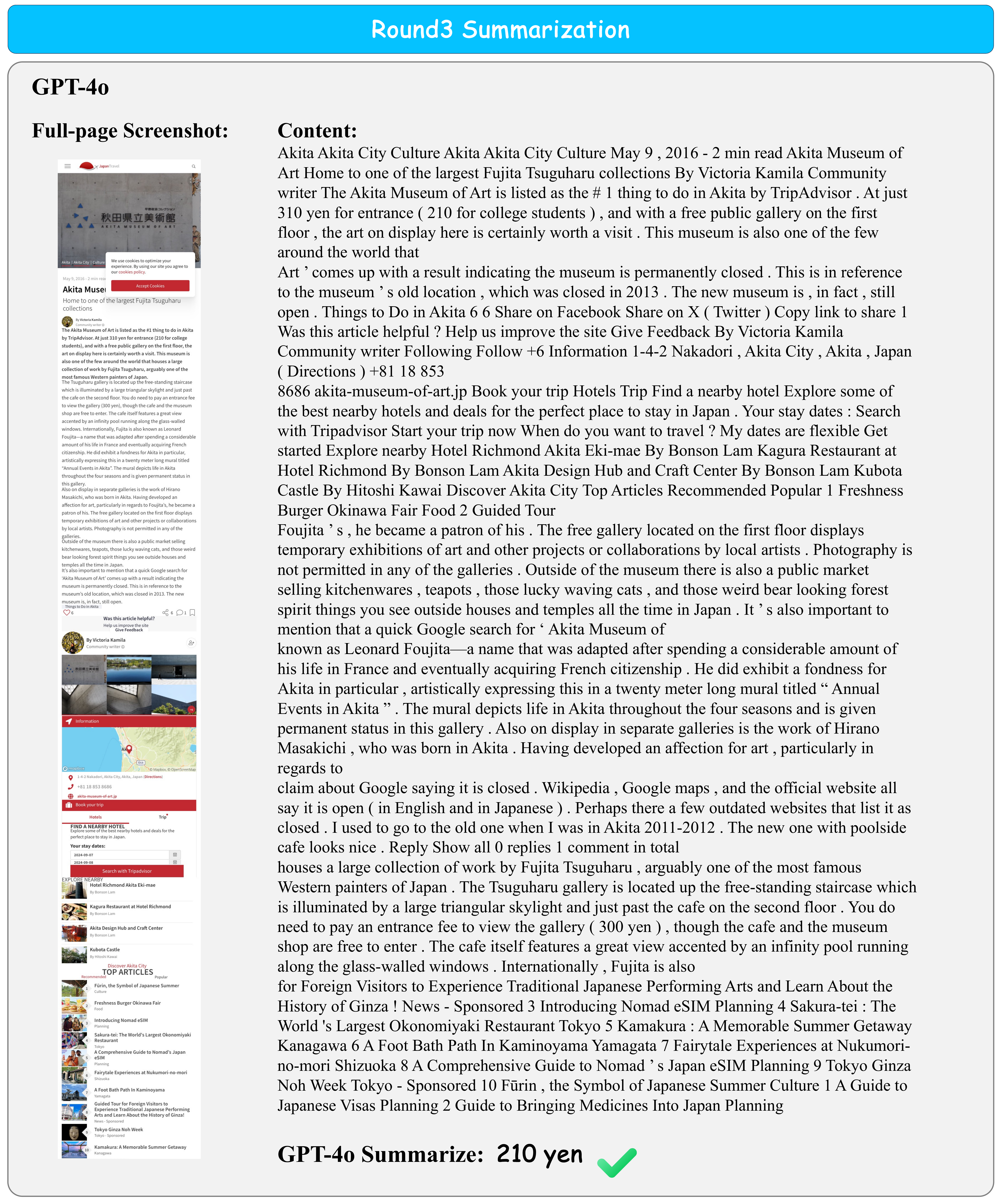} 
\caption{Response and middle results comparison of GPT-4o~\citep{openai2024gpt4o}, Qwen2-VL-7B~\citep{Qwen2-VL}, and LLaVA-OneVision-7B~\citep{li2024llava-ov} in the end-to-end task.}
\label{fig:data_sample_exp1_p1}
\end{figure}

\begin{figure}[ht]
\centering
\includegraphics[width=\textwidth]{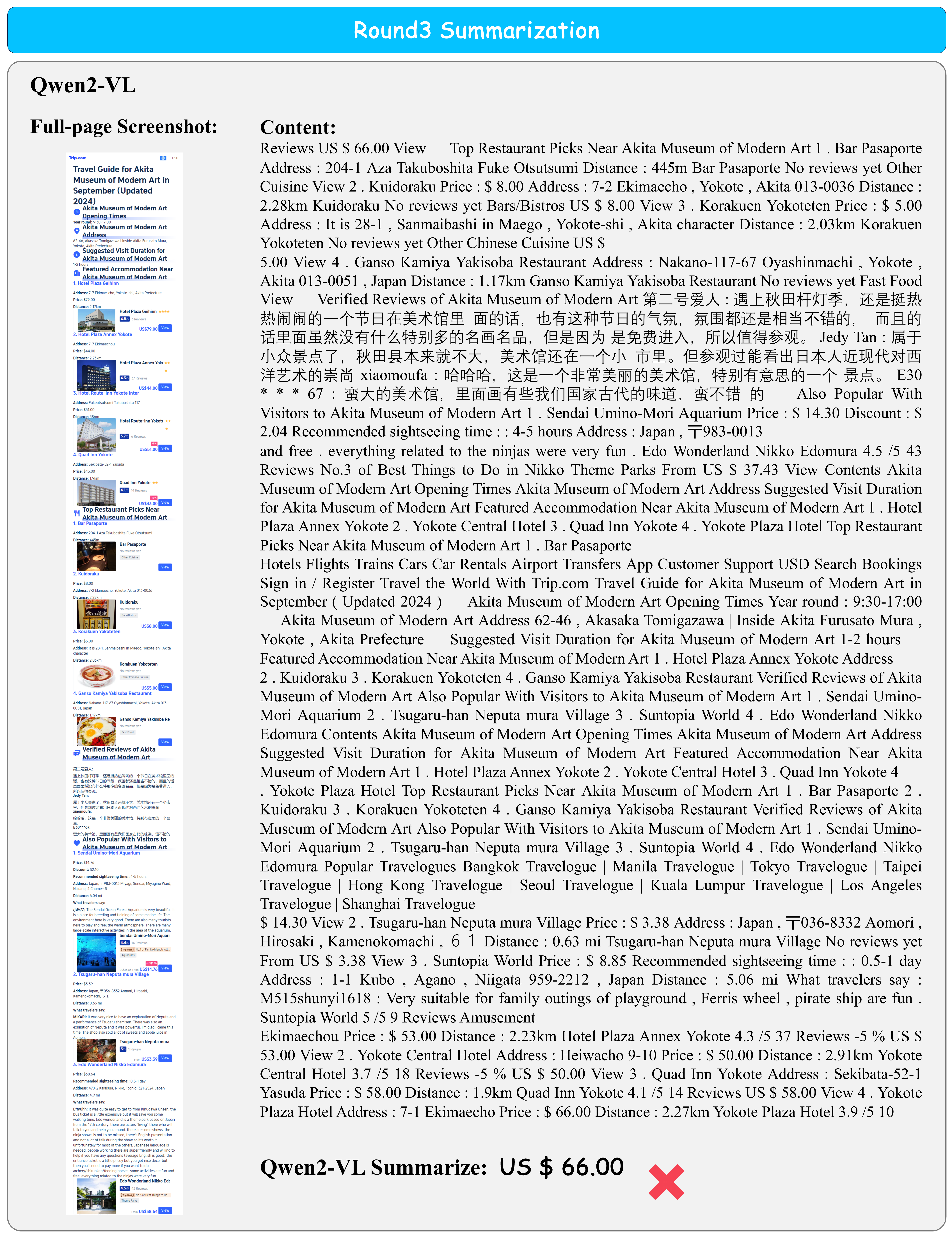} 
\caption{Response and middle results comparison of GPT-4o~\citep{openai2024gpt4o}, Qwen2-VL-7B~\citep{Qwen2-VL}, and LLaVA-OneVision-7B~\citep{li2024llava-ov} in the end-to-end task.}
\label{fig:data_sample_exp1_p1}
\end{figure}

\begin{figure}[ht]
\centering
\includegraphics[width=\textwidth]{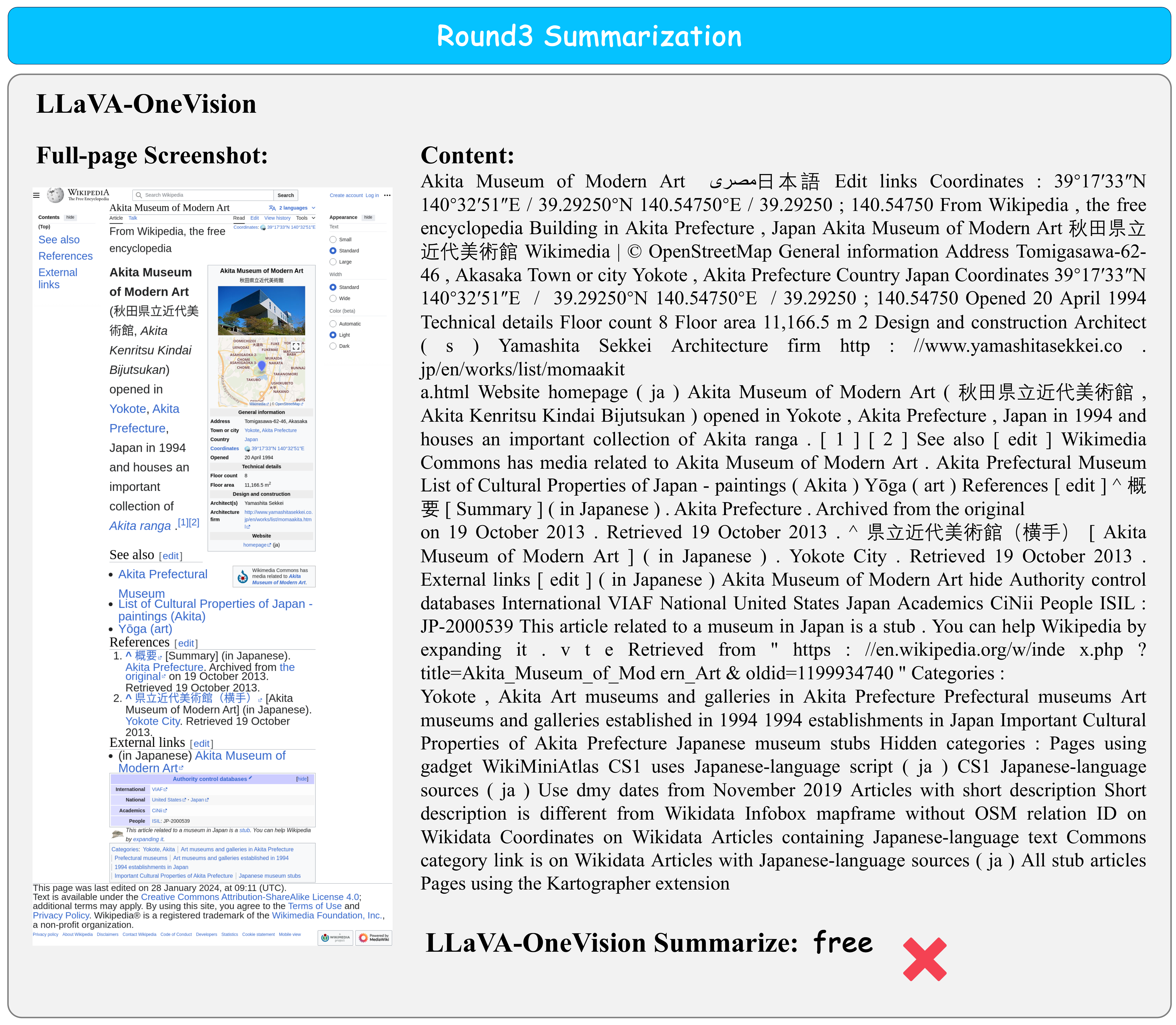} 
\caption{Response and middle results comparison of GPT-4o~\citep{openai2024gpt4o}, Qwen2-VL-7B~\citep{Qwen2-VL}, and LLaVA-OneVision-7B~\citep{li2024llava-ov} in the end-to-end task.}
\label{fig:data_sample_exp1_p1}
\end{figure}

\end{document}